\documentclass{article} 
\usepackage{iclr2017_conference,times}
\usepackage{hyperref}
\usepackage{url}

\usepackage{subfigure}
\usepackage{flafter}
\usepackage[pdftex]{graphicx}
\usepackage{tikz}
\usetikzlibrary{bayesnet}
\usepackage{amsfonts}
\usepackage{amsmath}
\usepackage{url}
\PassOptionsToPackage{hyphens}{url}\usepackage{hyperref}
\expandafter\def\expandafter\UrlBreaks\expandafter{\UrlBreaks
  \do\a\do\b\do\c\do\d\do\e\do\f\do\g\do\h\do\i\do\j%
  \do\k\do\l\do\m\do\n\do\o\do\p\do\q\do\r\do\s\do\t%
  \do\u\do\v\do\w\do\x\do\y\do\z\do\A\do\B\do\C\do\D%
  \do\E\do\F\do\G\do\H\do\I\do\J\do\K\do\L\do\M\do\N%
  \do\O\do\P\do\Q\do\R\do\S\do\T\do\U\do\V\do\W\do\X%
  \do\Y\do\Z}

\makeatletter
\newcommand{\figcaption}[1]{\def\@captype{figure}\caption{#1}}
\newcommand{\tblcaption}[1]{\def\@captype{table}\caption{#1}}
\makeatother

\newcommand{\cev}[1]{\reflectbox{\ensuremath{\vec{\reflectbox{\ensuremath{#1}}}}}}

\title{Neural Machine Translation with Latent Semantic of Image and Text}

\author{Joji Toyama$^{*}$, Masanori Misono\thanks{First two authors contributed equally.}\hspace{1.5mm}$^{\dagger}$, Masahiro Suzuki, Kotaro Nakayama \& Yutaka Matsuo\\
    Graduate School of Engineering, $^{\dagger}$Graduate School of Information Science and Technology\\
The University of Tokyo\\
Hongo, Tokyo, Japan  \\
\texttt{\{toyama,misono,masa,k-nakayama,matsuo\}@weblab.t.u-tokyo.ac.jp} \\
}

\begin{document}

\maketitle

\begin{abstract}
Although attention-based Neural Machine Translation have achieved great success, attention-mechanism cannot capture the entire meaning of the source sentence because the attention mechanism generates a target word depending heavily on the relevant parts of the source sentence. The report of earlier studies has introduced a latent variable to capture the entire meaning of sentence and achieved improvement on attention-based Neural Machine Translation. We follow this approach and we believe that the capturing meaning of sentence benefits from image information because human beings understand the meaning of language not only from textual information but also from perceptual information such as that gained from vision. As described herein, we propose a neural machine translation model that introduces a continuous latent variable containing an underlying semantic extracted from texts and images. Our model, which can be trained end-to-end, requires image information only when training. Experiments conducted with an English--German translation task show that our model outperforms over the baseline.
\end{abstract}

\section{Introduction}
\begin{figure}[t]
  \begin{center}
      \includegraphics[scale=0.3]{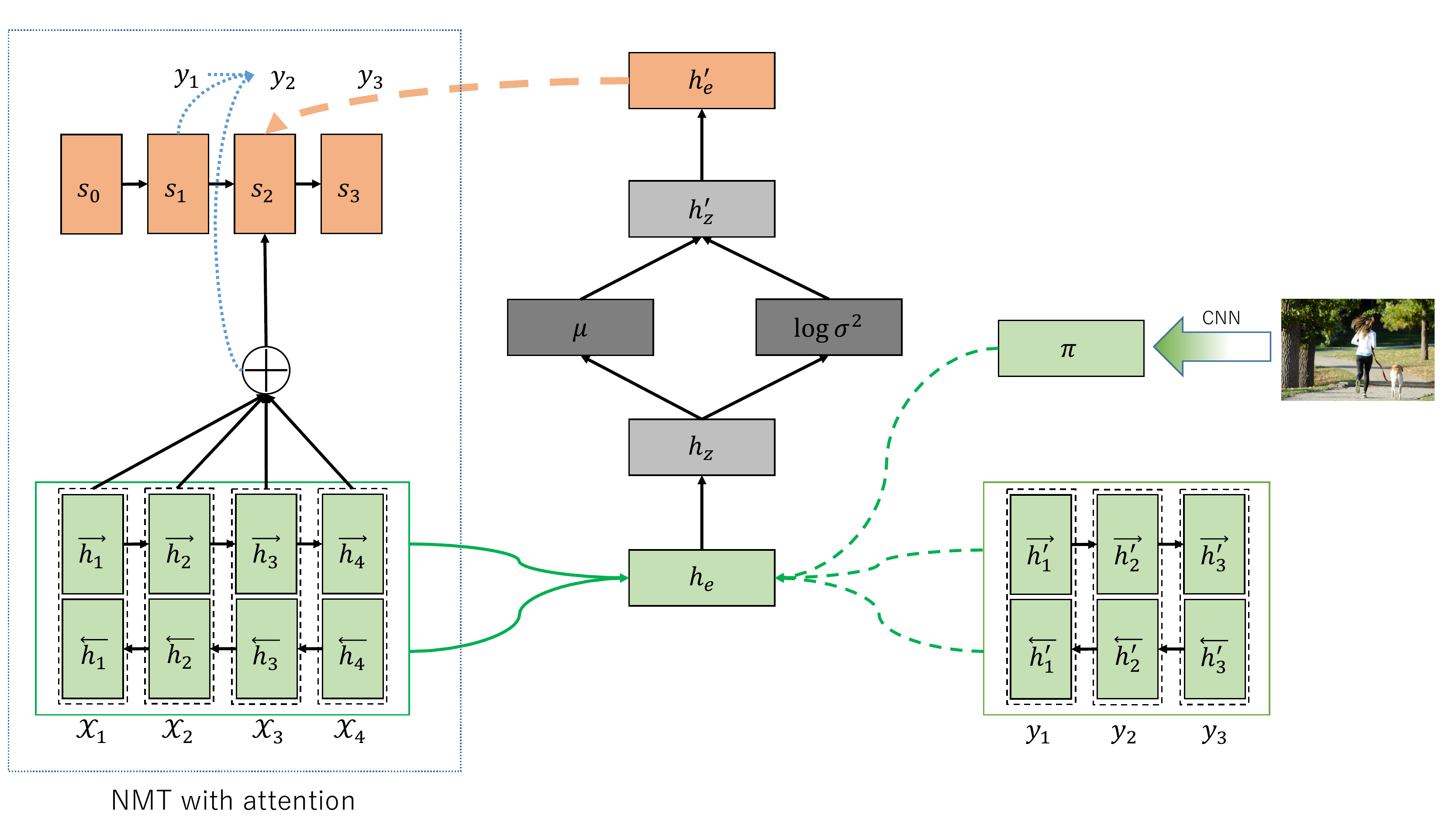}
    \caption{Architecture of Proposed Model.} Green dotted lines denote that $\boldsymbol {\pi}$ and encoded $\mathbf{y}$ are used only when training.
    \label{fig:mvnmt}
  \end{center}
\end{figure}
Neural machine translation (NMT) has achieved great success in recent years \citep{seq,nmt}. In contrast to statistical machine translation, which requires huge phrase and rule tables, NMT requires much less memory. However, the most standard model, NMT with attention \citep{nmt} entails the shortcoming that the attention mechanism cannot capture the entire meaning of a sentence because it generates a target word while depending heavily on the relevant parts of the source sentence \citep{tu}. To overcome this problem, Variational Neural Machine Translation (VNMT), which outperforms NMT with attention introduces a latent variable to capture the underlying semantic from source and target \citep{VNMT}. We follow  the motivation of VNMT, which is to capture underlying semantic of a source.

Image information is related to language. For example, we human beings understand the meaning of language by linking perceptual information given by the surrounding environment and language \citep{Barsalou1999}. Although it is natural and easy for humans, it is difficult for computers to understand different domain's information integrally. Solving this difficult task might, however, bring great improvements in natural language processing. Several researchers have attempted to link language and images such as image captioning by \citet{show} or image generation from sentences by \citet{gantext}. They described the possibility of integral understanding of images and text. In machine translation, we can expect an improvement using not only text information but also image information because image information can bridge two languages.

 As described herein, we propose the neural machine translation model which introduces a latent variable containing an underlying semantic extracted from texts and images. Our model includes an explicit latent variable $\mathbf{z}$, which has underlying semantics extracted from text and images by introducing a Variational Autoencoder (VAE) \citep{vae,vae2}. Our model, which can be trained end-to-end, requires image information only when training. As described herein, we tackle the task with which one uses a parallel corpus and images in training, while using a source corpus in translating. It is important to define the task in this manner because we rarely have a corresponding image when we want to translate a sentence. During translation, our model generates a semantic variable $\mathbf{z}$ from a source, integrates variable $\mathbf{z}$ into a decoder of neural machine translation system, and then finally generates the translation. The difference between our model and VNMT is that we use image information in addition to text information.

 For experiments, we used Multi30k \citep{multi30k}, which includes images and the corresponding parallel corpora of English and German. Our model outperforms the baseline with two evaluation metrics: METEOR \citep{meteor} and BLEU \citep{bleu}. Moreover, we obtain some knowledge related to our model and Multi30k. Finally, we present some examples in which our model either improved, or worsened, the result.
  
Our paper contributes to the neural machine translation research community in three ways.\\
\begin{itemize}
\item We present the first neural machine translation model to introduce a latent variable inferred from image and text information. We also present the first translation task with which one uses a parallel corpus and images in training, while using a source corpus in translating. 
\item Our translation model can generate more accurate translation by training with images, especially for short sentences.
\item We present how the translation of source is changed by adding image information compared to VNMT which does not use image information. 
\end{itemize}

\section{Background}
Our model is the extension of Variational Neural Machine Translation (VNMT) \citep{VNMT}. Our model is also viewed as one of the multimodal translation models. In our model, VAE is used to introduce a latent variable. We describe the background of our model in this section.
\subsection{Variational Neural Machine Translation}
    The VNMT translation model introduces a latent variable. This model's architecture shown in Figure \ref{fig:mvnmt} excludes the arrow from $\mathbf{\boldsymbol \pi}$. This model involves three parts: encoder, inferrer, and decoder. In the encoder, both the source and target are encoded by bidirectional-Recurrent Neural Networks (bidirectional-RNN) and a semantic representation is generated. In the inferrer, a latent variable $\mathbf{z}$ is modeled from a semantic representation by introducing VAE. In the decoder, a latent variable $\mathbf{z}$ is integrated in the Gated Recurrent Unit (GRU) decoder; also, a translation is generated. 
  
  Our model is followed by architecture, except that the image is also encoded to obtain a latent variable $\mathbf{z}$.
\subsection{Multimodal Translation}
Multimodal Translation is the task with which one might one can use a parallel corpus and images. The first papers to study multimodal translation are \citet{2015arXiv151004709E} and \citet{hitschler2016multimodal}. It was selected as a shared task in Workshop of Machine Translation 2016 (WMT16\footnote{http://www.statmt.org/wmt16/}). Although several studies have been conducted \citep{Does_multi,cmu,calixto-elliott-frank:2016:WMT,libovicky-EtAl:2016:WMT,rodriguezguasch,shah-wang}, they do not show great improvement, especially in neural machine translation \citep{result}. Here, we introduce end-to-end neural network translation models like our model.

  \citet{Does_multi} integrate an image into an NMT decoder. They simply put source context vectors and image feature vectors extracted from ResNet-50's `res4f\_relu' layer \citep{res} into the decoder called multimodal conditional GRU. They demonstrate that their method does not surpass the text-only baseline: NMT with attention.

  \citet{cmu} integrate an image into a head of source words sequence. They extract prominent objects from the image by Region-based Convolutional Neural Networks (R-CNN) \citep{frcnn}. Objects are then converted to feature vectors by VGG-19 \citep{VGG} and are put into a head of source words sequence. They demonstrate that object extraction by R-CNN contributes greatly to the improvement. This model achieved the highest METEOR score in NMT-based models in WMT16, which we compare to our model in the experiment. We designate this model as CMU.

\citet{Does_multi} argue that their proposed model did not achieve improvement because they failed to benefit from both text and images. We assume that they failed to integrate text and images because they simply put images and text into neural machine translation despite huge gap exists between image information and text information. Our model, however, presents the possibility of benefitting from images and text because text and images are projected to their common semantic space so that the gap of images and text would be filled.

\subsection{Variational Auto Encoder}
  VAE was proposed in an earlier report of the literature \citet{vae,vae2}. Given an observed variable $\mathbf{x}$, VAE introduces a continuous latent variable $\mathbf{z}$, with the assumption that $\mathbf{x}$ is generated from $\mathbf{z}$. VAE incorporates $p_\theta(\mathbf{x|z})$ and $q_\phi(\mathbf{z|x})$ into an end-to-end neural network. The lower bound is shown below.
\begin{equation}
    \mathcal{L}_{\mathrm{VAE}} = -\mathrm{D}_{\mathrm{KL}}\left[q_\phi(\mathbf{z|x}) || p_\theta(\mathbf{z}) \right] + \mathbb{E}_{q_\phi(\mathbf{z|x})}\left[ \log p_\theta(\mathbf{x|z}) \right] \leq \log p_\theta(\mathbf{x})
\end{equation}

\section{Neural Machine Translation With Latent Semantic Of Image And Text}
We propose a neural machine translation model which explicitly has a latent variable containing an underlying semantic extracted from both text and image. This model can be seen as an extension of VNMT by adding image information.

Our model can be drawn as a graphical model in Figure \ref{fig:type1}. Its lower bound is
\begin{equation}
\mathcal{L} = -\mathrm{D}_{\mathrm{KL}}\left[q_\phi(\mathbf{z|x,y,\boldsymbol \pi}) || p_\theta(\mathbf{z|x}) \right] + \mathbb{E}_{q_\phi(\mathbf{z|x,y,\boldsymbol \pi})}\left[ \log p_\theta(\mathbf{y|z,x}) \right],
\label{eq:lower1}
\end{equation}
\begin{table}
    \begin{center}
        \begin{tabular}{cc}
        	    \begin{minipage}{0.46\hsize}
        	    \begin{center}
        	        \begin{tabular}{cc}
            \begin{tikzpicture}
                \node[latent]                               (z) {$z$};
                \node[obs, below=of z, xshift=-1.0cm] (x) {$\mathbf{x}$};
                \node[obs, below=of z, xshift=1.0cm]  (y) {$\mathbf{y}$};

                \edge {x} {z} ; %
                \edge {z} {y}
                \edge {x} {y} ; %

            \end{tikzpicture} \hspace{0.5cm}&

            \begin{tikzpicture}

                \node[latent]                               (z) {$z$};
                \node[obs, below=of z, xshift=-1.0cm] (x) {$\mathbf{x}$};
                \node[obs, below=of z, xshift=1.0cm]  (y) {$\mathbf{y}$};

                \dedge {x,y} {z} ;
            \end{tikzpicture}
        \end{tabular}
        \figcaption{VNMT}
        \label{fig:vnmt}
    \end{center}
    \end{minipage}
    \begin{minipage}{0.46\hsize}
    \begin{center}
        \begin{tabular}{cc}
            \begin{tikzpicture}
                \node[latent]                               (z) {$z$};
                \node[obs, below=of z, xshift=-1.2cm] (x) {$\mathbf{x}$};
                \node[obs, below=of z, xshift=1.2cm]  (y) {$\mathbf{y}$};

                \edge {x} {z} ; %
                \edge {z} {y}
                \edge {x} {y} ; %

            \end{tikzpicture} \hspace{1cm}&

            \begin{tikzpicture}

                \node[latent]                               (z) {$z$};
                \node[obs, below=of z, xshift=-1.2cm] (x) {$\mathbf{x}$};
                \node[obs, below=of z, xshift=1.2cm]  (y) {$\mathbf{y}$};
                \node[obs, above=of z]            (pi) {$\pi$};

                \dedge {x,y, pi} {z} ;
            \end{tikzpicture}
        \end{tabular}
        \figcaption{Our model}
        \label{fig:type1}

    \end{center}
    \end{minipage}
    \end{tabular}
    \end{center}
\end{table}
where $\mathbf{x,y,\boldsymbol \pi,z}$ respectively denote the source, target, image and latent variable, and  $p_\theta$ and $q_\phi$ respectively denote the prior distribution and the approximate posterior distribution. It is noteworthy in Eq. (\ref{eq:lower1}) that we want to model $p(\mathbf{z|x,y,\boldsymbol \pi})$, which is intractable. Therefore we model $q_\phi(\mathbf{z|x,y,\boldsymbol \pi})$ instead, and also model prior $p_\theta(\mathbf{z|x})$ so that we can generate a translation from the source in testing. Derivation of the formula is presented in the appendix. 
  
We model all distributions in Eq. (\ref{eq:lower1}) by neural networks. Our model architecture is divisible into three parts:  1) encoder, 2) inferrer, and 3) decoder.

\subsection{Encoder}
In the encoder, the semantic representation $\mathbf{h_e}$ is obtained from the image, source, and target. We propose several methods to encode an image. We show how these methods affect the translation result in the Experiment section. This representation is used in the inferrer. This section links to the green part of Figure \ref{fig:mvnmt}.

\subsubsection{text encoding}
The source and target are encoded in the same way as \cite{nmt}. The source is converted to a sequence of 1-of-k vector and is embedded to $d_{emb}$ dimensions. We designate it as the source sequence.
Then, a source sequence is put into bidirectional RNN. Representation $\mathbf{h}_i$ is obtained by concatenating $\vec{\mathbf{h}}_i$ and $\cev{\mathbf{h}}_i$ : $\vec{\mathbf{h}}_i = \mathrm{RNN}(\vec{\mathbf{h}}_{i-1},E_{w_i}), \cev{\mathbf{h}}_i = \mathrm{RNN}({\cev{\mathbf{h}}}_{i+1},E_{w_i}), \mathbf{h}_i = [\vec{\mathbf{h}}_i; \cev{\mathbf{h}}_i]$, where $E_{w_i}$ is the embedded word in a source sentence, $\mathbf{h}_i \in \mathbb{R}^{d_h}$, and $\vec{\mathbf{h}}_i, \cev{\mathbf{h}}_i \in \mathbb{R}^{\frac{d_h}{2}}$. It is conducted through $i = 0$ to $i = T_f$, where $T_f$ is the sequence length. GRU is implemented in bidirectional RNN so that it can attain long-term dependence. Finally, we conduct mean-pooling to $\mathbf{h}_i$ and obtain the source representation vector as
$\mathbf{h}_f=\frac{1}{T_f} \sum^{T_f}_i \mathbf{h}_i$.
 The exact same process is applied to target to obtain target representation $\mathbf{h}_g$.

\subsubsection{image encoding and semantic representation}
We use Convolutional Neural Networks (CNN) to extract feature vectors from images. We propose several ways of extracting image features.

\begin{description}
    \item[Global (G)] The image feature vector is extracted from the image using a CNN. With this method, we use a feature vector in the certain layer as $\boldsymbol \pi$. Then $\boldsymbol \pi$ is encoded to the image representation vector $\mathbf{h}_\pi$ simply by affine transformation as
\begin{equation}
\mathbf{h}_\pi=W_\pi {\boldsymbol \pi} + b_\pi \quad \mbox{where} \; W_\pi \in \mathbb{R} ^ {d_\pi \times d_{fc7}} \; , \; b_\pi \in \mathbb{R} ^ {d_\pi}.
\label{affine}
\end{equation}

\item[Global and Objects (G+O)] First we extract some prominent objects from images in some way. Then, we obtain fc7 image feature vectors $\boldsymbol \pi$ from the original image and extracted objects using a CNN. Therefore $\boldsymbol \pi$ takes a variable length. We handle $\boldsymbol \pi$ in two ways: average and RNN encoder.

 In average ({\bf G+O-AVG}), we first obtain intermediate image representation vector $\mathbf{h}'_\pi$ by affine transformation in Eq. (\ref{affine}). Then, the average of $\mathbf{h}'_\pi$ becomes the image representation vector: $\mathbf{h}_\pi = \frac{\sum^l_i \mathbf{h}'_{\pi_i}}{l}$, where $l$ is the length of $\mathbf{h}'_\pi$. 
 
 In RNN encoder ({\bf G+O-RNN}), we first obtain $\mathbf{h}'_\pi$ by affine transformation in Eq. (\ref{affine}). Then, we encode $\mathbf{h}'_\pi$ in the same way as we encode text in Section 3.1.1 to obtain $\mathbf{h}_\pi$. 
 
\item[Global and Objects into source and target (G+O-TXT)] Thereby, we first obtain $\mathbf{h}'_\pi$ by affine transformation in Eq. (\ref{affine}). Then, we put sequential vector $\mathbf{h}'_\pi$ into the head of the source sequence and target sequence. In this case, we set $d_\pi$ to be the same dimension as $d_{emb}$. In fact, the source sequence including $\mathbf{h}'_\pi$ is only used to model $q_\phi(\mathbf{z|x,y,\boldsymbol \pi})$. Context vector $\mathbf{c}$ (Eq. (\ref{eq:context})) and $p_\theta(\mathbf{z|x})$ are computed by a source sequence that does not include $\mathbf{h}'_\pi$. We encode the source sequence including $\mathbf{h}'_\pi$ as Section 3.1.1 to obtain $\mathbf{h}_f$ and $\mathbf{h}_g$. In this case, $\mathbf{h}_\pi$ is not obtained. Image information is contained in $\mathbf{h}_f$ and $\mathbf{h}_g$.

\end{description}

  All representation vectors $\mathbf{h}_f$, $\mathbf{h}_g$ and $\mathbf{h}_\pi$ are concatenated to obtain a semantic representation vector as $\mathbf{h}_e = [\mathbf{h}_f ; \mathbf{h}_g ; \mathbf{h}_\pi]$, where $\mathbf{h}_e \in \mathbb{R}^{d_e = 2 \times d_h + d_\pi}$ (in G+O-TXT: $\mathbf{h}_e = [\mathbf{h}_f ; \mathbf{h}_g]$, where $\mathbf{h}_e \in \mathbb{R}^{d_e = 2 \times d_h}$). It is an input of the multimodal variational neural inferrer.
  
\subsection{Inferrer}
We model the posterior $q_\phi(\mathbf{z|x,y,\boldsymbol \pi})$ using a neural network and also the prior $p_\theta(\mathbf{z|x})$ by neural network. This section links to the black and grey part of Figure \ref{fig:mvnmt}.

\subsubsection{Neural Posterior Approximator}
\label{ssec:neural_posterior_approximator}
Modeling the true posterior $p_\theta(\mathbf{z|x,y,\boldsymbol \pi})$ is usually intractable. Therefore, we consider modeling of an approximate posterior $q_\phi(\mathbf{z|x,y,\boldsymbol \pi})$ by introducing VAE. We assume that the posterior $q_\phi(\mathbf{z| x,y,\boldsymbol \pi})$ has the following form:
\begin{equation}
    \label{eq:ga}
    q_\phi(\mathbf{z|x,y,\boldsymbol \pi})= \mathcal{N}(\mathbf{z;\boldsymbol \mu(x,y,\boldsymbol \pi),\boldsymbol \sigma(x,y,\boldsymbol \pi})^2\mathbf{I}).
 \end{equation}
The mean $\boldsymbol \mu$ and standard deviation $\boldsymbol \sigma$ of the approximate posterior are the outputs of neural networks.

Starting from the variational neural encoder, a semantic representation vector $\mathbf{h}_e$ is projected to latent semantic space as
\begin{equation}
    \mathbf{h}_z = g(W^{(1)}_z \mathbf{h}_e+ \mathbf{b}^{(1)}_z),
\label{eq:hz}
\end{equation}
where $W_z^{(1)} \in \mathbb{R} ^ {d_z \times (d_e)} \; \; \mathbf{b}_z^{(1)} \in \mathbb{R} ^ {d_z}$.
$g(\cdot)$ is an element-wise activation function, which we set as $\mathrm{tanh}(\cdot)$.
Gaussian parameters of Eq. (\ref{eq:ga}) are obtained through linear regression as
\begin{equation}
    {\boldsymbol \mu} = W_\mu \mathbf{h}_z + \mathbf{b}_\mu, \log {\boldsymbol \sigma}^2 = W_\sigma \mathbf{h}_z + \mathbf{b}_\sigma,
\end{equation}
where $\; {\boldsymbol \mu}, \log {\boldsymbol \sigma}^2 \in  \mathbb{R} ^ {d_z}$.

\subsubsection{Neural Prior Model}
We model the prior distribution $p_\theta(\mathbf{z|x})$ as follows:
\begin{eqnarray}
    p_\theta(\mathbf{z|x}) &=&  \mathcal{N}(\mathbf{z;\boldsymbol \mu'(x),\boldsymbol \sigma'(x)}^2\mathbf{I}) \nonumber. \\
\end{eqnarray}
$\mathbf{\boldsymbol \mu'}$ and $\mathbf{\boldsymbol \sigma'}$ are generated in the same way as that presented in Section \ref{ssec:neural_posterior_approximator}, except for the absence of $\mathbf{y}$ and $\boldsymbol \pi$ as inputs. Because of the absence of representation vectors, the dimensions of weight in equation (\ref{eq:hz}) for prior model are
$W_z^{'(1)} \in \mathbb{R} ^ {d_z \times d_h}, \; \mathbf{b}_z^{'(1)} \in \mathbb{R} ^ {d_z}$.
  We use a reparameterization trick to obtain a representation of latent variable $\mathbf{z}$: 
  $\mathbf{h}'_z = \boldsymbol \mu + \boldsymbol \sigma  \boldsymbol \epsilon$, $\boldsymbol \epsilon \sim \mathcal{N}(0,I)$. During translation, $\mathbf{h}'_z$ is set as the mean of $p_\theta(\mathbf{z|x})$.
  Then, $\mathbf{h}'_z$ is projected onto the target space as
  \begin{equation}
      \mathbf{h}'_e =  g(W^{(2)}_z \mathbf{h}'_z + \mathbf{b}^{(2)}_z) \quad \mbox{where} \; \mathbf{h}'_e \in \mathbb{R} ^ {d_e}.
  \end{equation}
  $\mathbf{h}'_e$ is then integrated into the neural machine translation's decoder.
  
\subsection{Decoder}
This section links to the orange part of Figure \ref{fig:mvnmt}.
Given the source sentence $\mathbf{x}$ and the latent variable $\mathbf{z}$, decoder defines the probability over translation $\mathbf{y}$ as
\begin{eqnarray}
p(\mathbf{y|z,x}) = \prod_{j=1}^T p(\mathbf{y}_j | \mathbf{y}_{<j}, \mathbf{z, x}).
\end{eqnarray}
How we define the probability over translation $\mathbf{y}$ is fundamentally the same as VNMT, except for using conditional GRU instead of GRU. Conditional GRU involves two GRUs and an attention mechanism. We integrate a latent variable $\mathbf{z}$ into the second GRU. We describe it in the appendix.  
\subsection{Model Training}
Monte Carlo sampling method is used to approximate the expectation over the posterior Eq. (\ref{eq:lower1}), $\mathbb{E}_{q_\phi(\mathbf{z|x,y,\boldsymbol \pi})} \approx \frac{1}{L} \sum_{l=1}^L \log p_\theta(\mathbf{y | x,h}^{(l)}_z)$, where $L$ is the number of samplings. The training objective is defined as
\begin{eqnarray}
\mathcal{L}(\theta,\phi) = -\mathrm{D}_{\mathrm{KL}}\left[q_\phi(\mathbf{z|x,y,\boldsymbol \pi}) || p_\theta(\mathbf{z|x}) \right] + \frac{1}{L} \sum_{l=1}^L \sum_{j=1}^T \log p_\theta(\mathbf{y}_j | \mathbf{y}_{< j},\mathbf{x,h}_z^{(l)}),
\end{eqnarray}
where $\mathbf{h}_z = {\boldsymbol \mu + \boldsymbol \sigma \cdot \boldsymbol \epsilon}$, ${\boldsymbol \epsilon} \sim \mathcal{N} (0,I)$. The first term, KL divergence, can be computed analytically and is differentiable because both distributions are Gaussian. The second term is also differentiable. We set $L$ as 1. Overall, the objective $\mathcal{L}$ is differentiable. Therefore, we can optimize the parameter $\theta$ and variational parameter $\phi$ using gradient ascent techniques.

\section{Experiments}
\subsection{Experimental Setup}
We used Multi30k \citep{multi30k} as the dataset. Multi30k have an English description and a German description for each corresponding image. We handle 29,000 pairs as training data, 1,014 pairs as validation data, and 1,000 pairs as test data.

 Before training, punctuation normalization and lowercase are applied to both English and German sentences by Moses \citep{moses} scripts\footnote{https://github.com/moses-smt/mosesdecoder/blob/master/scripts/tokenizer/\{normalize-punctuation, lowercase, tokenizer, detokenizer\}.perl}.
 Compound-word splitting is conducted only to German sentences using \citet{subword-nmt}\footnote{https://github.com/rsennrich/subword-nmt}. Then we tokenize sentences\footnotemark[2] and use them as training data. We produce vocabulary dictionaries from training data. The vocabulary becomes 10,211 words for English and 13,180 words for German after compound-word splitting. 

Image features are extracted using VGG-19 CNN \citep{VGG}. We use 4096-dimensional fc7 features.
To extract the object's region, we use Fast R-CNN \citep{frcnn}. Fast R-CNN is trained on ImageNet and MSCOCO dataset \footnote{https://github.com/rbgirshick/fast-rcnn/tree/coco}.


All weights are initialized by $\mathcal{N}(0,0.01\mathbf{I})$. We use the adadelta algorithm as an optimization method. 
The hyperparameters used in the experiment are presented in the Appendix. All models are trained with early stopping.
When training, VNMT is fine-tuned by NMT model and our models are fine-tuned using VNMT. 
When translating, we use beam-search. The beam-size is set as 12.
Before evaluation, we restore split words to the original state and de-tokenize\footnotemark[2] generated sentences.

We implemented proposed models based on {\it dl4mt}\footnote{https://github.com/nyu-dl/dl4mt-tutorial}. Actually, {\it dl4mt} is fundamentally the same model as \citet{nmt}, except that its decoder employs conditional GRU\footnote{The architecture is described at https://github.com/nyu-dl/dl4mt-tutorial/blob/master/docs/cgru.pdf}. We implemented VNMT also with conditional GRU so small difference exists between our implementation and originally proposed VNMT which employs normal GRU as a decoder.  We evaluated results based on METEOR and BLUE using {\tt MultEval}
\footnote{https://github.com/jhclark/multeval, we use meteor1.5 instead of meteor1.4, which is the default of {\tt MultEval}. }.

\subsection{Result}
\begin{figure}[t]
\begin{tabular}{c}
\begin{minipage}{0.46\hsize}
  \begin{center}
      \includegraphics[scale=0.33]{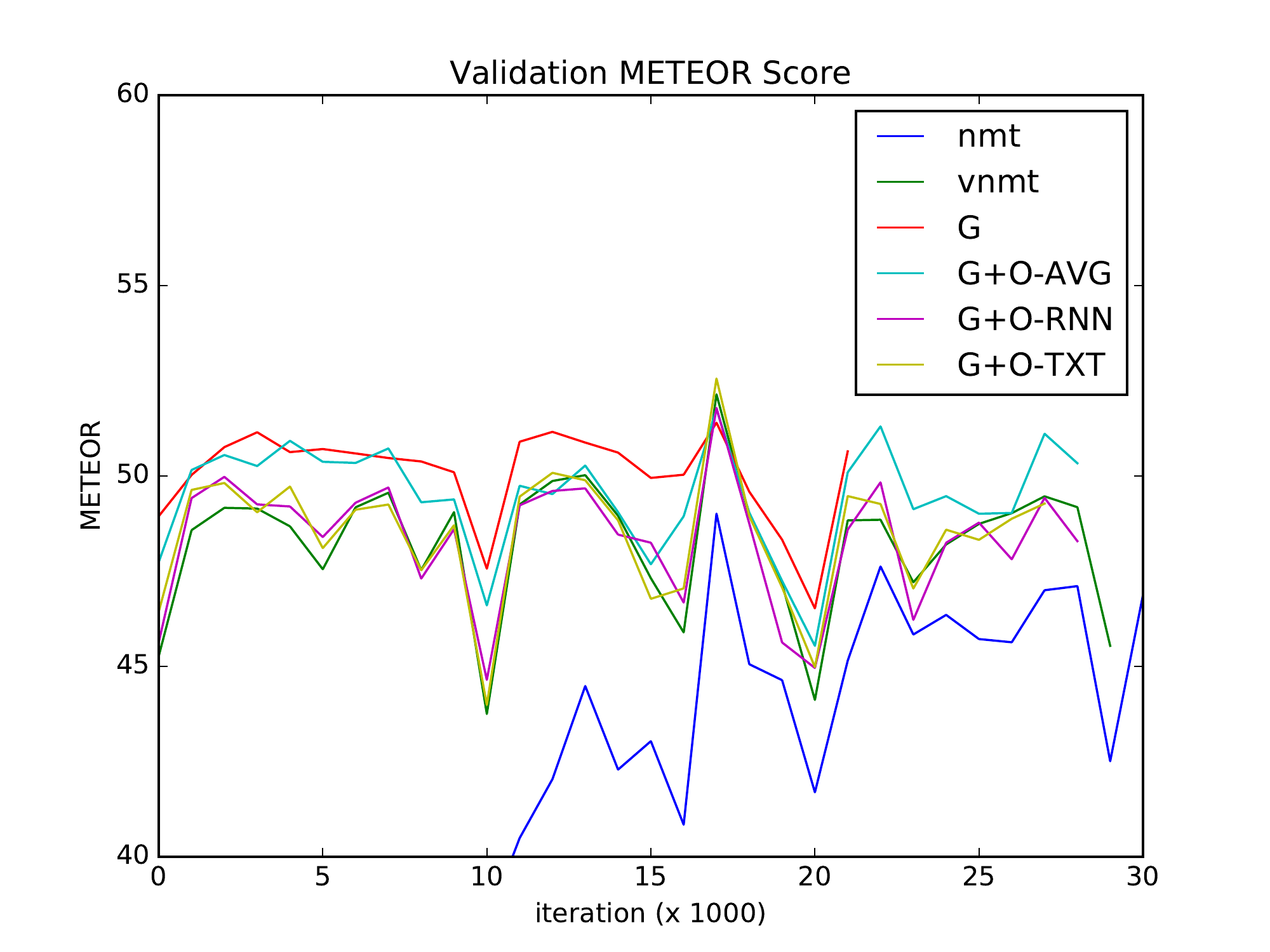}
      \caption{METEOR score to the validation data which are calculated for each 1000 iterations.}
    \label{fig:meteor}
  \end{center}
\end{minipage}
\hspace{10mm}

\begin{minipage}{0.46\hsize}
  \begin{center}
      \includegraphics[scale=0.33]{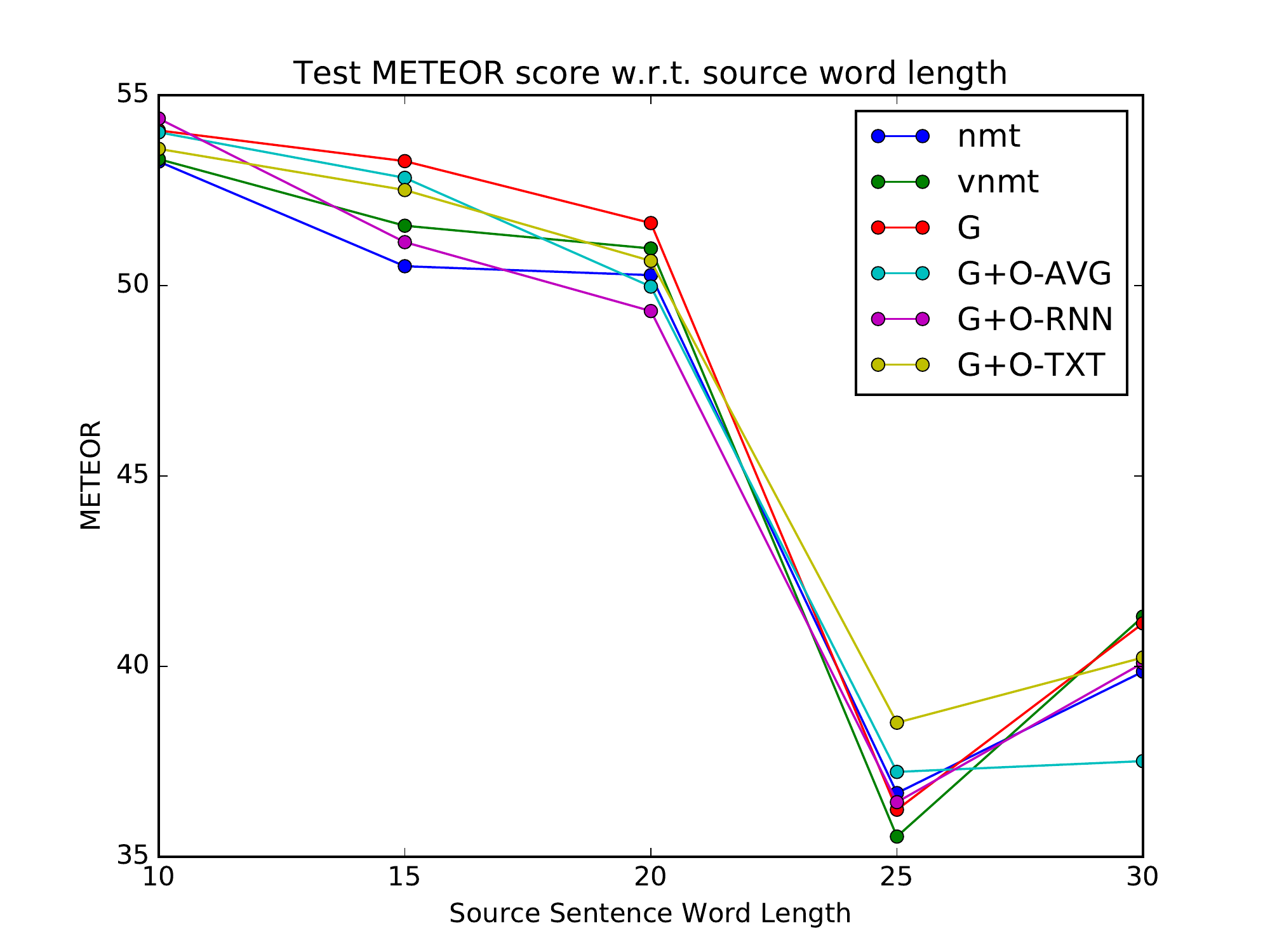}
    \caption{METEOR score on different groups of the source sentence length.}
    \label{fig:length}
  \end{center}
\end{minipage}
\end{tabular}
\end{figure}

Table \ref{table:result} presents experiment results. It shows that our models outperforms the baseline in both METEOR and BLEU. Figure \ref{fig:meteor} shows the plot of METEOR score of baselines and our models models in validation. Figure \ref{fig:length} shows the plot of METEOR score and the source sentence length.

\begin{table}[htbp]
    \begin{center}
        \caption{Evaluation Result on Multi30k dataset (English--German).
        The scores in parentheses are computed with `-norm' parameter. NMT is {\it dl4mt}'s NMT (in the {\it session3 } directory). The score of the CMU is from \citep{cmu}. }
        \vspace{2mm}
        \label{table:result}
        \begin{tabular}{ll|ll|ll}
                        & & \multicolumn{2}{|c|}{METEOR $\uparrow$} & \multicolumn{2}{|c}{BLEU $\uparrow$} \\ \hline
                        & & \multicolumn{1}{c}{val} & \multicolumn{1}{c}{test}  & \multicolumn{1}{|c}{val} & \multicolumn{1}{c}{test}\\
                        \hline
                        & NMT             & 51.5       (55.8)  & 50.5 (54.9)         &35.8       &   33.1    \\
                        & VNMT            & 52.2 (56.3)  & 51.1 (55.3)         &{\bf 37.0} &   34.9    \\
                        & CMU            &  \ \ \ - \ \ \   (-) & \ \  - \ \   \ \ (54.1)     & \ \ -  & \ \ -    \\
            \hline
            Our Model & G             & 50.6       (54.8)  & {\bf 52.4 (56.0)}   &34.5       &   {\bf 36.5}    \\
                        & G+O-AVG        & 51.8       (55.8)  & 51.8 (55.8)         &35.7       &   35.8    \\
                        & G+O-RNN        & 51.8       (56.1)  & 51.0 (55.4)         &35.9       &   34.9    \\
                        & G+O-TXT        & \bf{52.6       (56.8)}  & 51.7 (56.0)         &36.6       &   35.1    \\
        \end{tabular}
    \end{center}
\end{table}

\subsection{Quantitative Analysis}
Table \ref{table:result} shows that G scores the best in proposed models. In G, we simply put the feature of the original image. Actually, proposed model does not benefit from R-CNN, presumably because we can not handle sequences of image features very well. For example, G+O-AVG uses the average of multiple image features, but it only makes the original image information unnecessarily confusing.

Figure \ref{fig:meteor} shows that G and G+O-AVG outperforms VNMT almost every time, but all model scores increase suddenly in the 17,000 iteration validation. We have no explanation for this behavior. Figure \ref{fig:meteor} also shows that G and G+O-AVG scores fluctuate more moderately than others. We state that G and G+O-AVG gain stability by adding image information. When one observes the difference between the test score and the validation score for each model, baseline scores decrease more than proposed model scores. Especially, the G score increases in the test, simply because proposed models produce a better METEOR score on average, as shown in Figure \ref{fig:meteor}.

 Figure \ref{fig:length} shows that G and G+O-AVG make more improvements on baselines in short sentences than in long sentences, presumably because $q_\phi(\mathbf{z|x,y,\boldsymbol \pi})$ can model $\mathbf{z}$ well when a sentence is short. Image features always have the same dimension, but underlying semantics of the image and text differ. We infer that when the sentence is short, image feature representation can afford to approximate the underlying semantic, but when a sentence is long, image feature representation can not approximate the underlying semantic.

 Multi30k easily becomes overfitted, as shown in Figure \ref{fig:nmt_cost_compare} and \ref{fig:nmt_meteor_compare} in the appendix. This is presumably because 1) Multi30k is the descriptions of image, making the sentences short and simple, and 2) Multi30k has 29,000 sentences, which could be insufficient. In the appendix, we show how the parameter setting affects the score. One can see that decay-c has a strong effect. \citet{cmu} states that their proposed model outperforms the baseline (NMT), but we do not have that observation. It can be assumed that their baseline parameters are not well tuned.

\subsection{Qualitative Analysis}
 We presented the top 30 sentences, which make the largest METEOR score difference between G and VNMT, to native speakers of German and get the overall comments. They were not informed of our model training with image in addition to text. These comments are summarized into two general remarks. One is that G translates the meaning of the source material more accurately than VNMT. The other is that our model has more grammatical errors as prepositions' mistakes or missing verbs compared to VNMT. We assume these two remarks are reasonable because G is trained with images which mainly have a representation of noun rather than verb, therefore can capture the meaning of materials in sentence.
  
  Figure \ref{fig:trans1} presents the translation results and the corresponding image which G translates more accurately than VNMT in METEOR. Figure \ref{fig:trans2} presents the translation results and the corresponding image which G translates less accurately than VNMT in METEOR. Again, we note that our model does not use image during translating. In Figure \ref{fig:trans1}, G translates "a white and black dog" correctly while VNMT translates it incorrectly implying "a white dog and a black dog". We assume that G correctly translates the source because G captures the meaning of material in the source. In Figure \ref{fig:trans2}, G incorrectly translates the source. Its translation result is missing the preposition meaning "at", which is hardly represented in image.We present more translation examples in appendix.

\begin{table}[htbp]
    \begin{center}
        \begin{tabular}{|l|l|}
            \multicolumn{2}{c}{\includegraphics[scale=0.5]{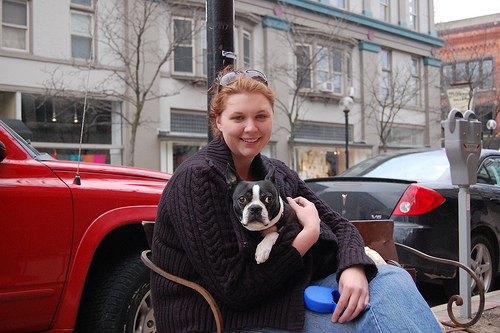}}\\
            \hline
            {\bf Source}  &
            a woman holding a white and black dog.
            \\ \hline
            {\bf Truth}  &
            eine frau h\"alt einen wei\ss-schwarzen hund.
            \\ \hline
            {\bf VNMT} &
            eine frau h\"alt einen wei\ss en und schwarzen hund.
            \\ \hline
            {\bf Our Model (G)} &
            eine frau h\"alt einen wei\ss-schwarzen hund.
            \\ \hline
        \end{tabular}
        \figcaption{Translation 1}
        \label{fig:trans1}
    \end{center}
\end{table}


\begin{table}[htbp]
    \begin{center}
        \begin{tabular}{|l|l|}
            \multicolumn{2}{c}{\includegraphics[scale=0.5]{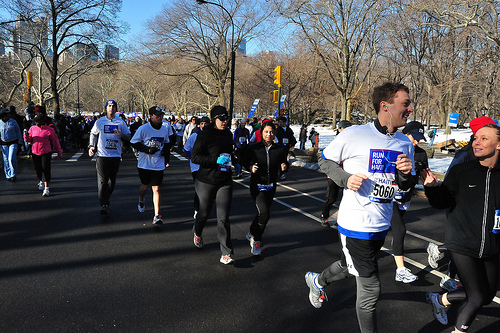}}\\
            \hline
            {\bf Source}  &
            a group of people running a marathon in the winter.
            \\ \hline
            {\bf Truth}  &
            eine gruppe von menschen l\"auft bei einem marathon im winter.
            \\ \hline
            {\bf VNMT} &
            eine gruppe von menschen l\"auft bei einem marathon im winter.
            \\ \hline
            {\bf Our Model (G)} &
            eine gruppe leute l\"auft einen marathon im winter an.
            \\ \hline
        \end{tabular}
        \figcaption{Translation 2}
        \label{fig:trans2}
    \end{center}
\end{table}

\section{Conclusion}
As described herein, we proposed the neural machine translation model that explicitly has a latent variable that includes underlying semantics extracted from both text and images. 
Our model outperforms the baseline in both METEOR and BLEU scores. Experiments and analysis present that our model can generate more accurate translation for short sentences. In qualitative analysis, we present that our model can translate nouns accurately while our model make grammatical errors. 


\bibliography{iclr2017_conference}
\bibliographystyle{iclr2017_conference}

\newpage

\appendix
\section{Derivation of Lower bounds}
The lower bound of our model can be derived as follows:
\begin{eqnarray*}
    p(\mathbf{y}|\mathbf{x}) &=& \int p(\mathbf{y},\mathbf{z} | \mathbf{x}) d\mathbf{z}  \\
                             &=& \int p(\mathbf{z}|\mathbf{x}) p(\mathbf{y}|\mathbf{z},\mathbf{x}) d\mathbf{z}
\end{eqnarray*}

\begin{eqnarray*}
\log p(\mathbf{y}|\mathbf{x}) &=& \log \int q(\mathbf{z}|\mathbf{x},\mathbf{y},\mathbf{\boldsymbol \pi}) \frac{p(\mathbf{z}|\mathbf{x}) p(\mathbf{y}|\mathbf{z},\mathbf{x})}{q(\mathbf{z}|\mathbf{x},\mathbf{y},\mathbf{\boldsymbol \pi})} d\mathbf{z} \\
&\geq& \int q(\mathbf{z}|\mathbf{x},\mathbf{y},\mathbf{\boldsymbol \pi}) \log \frac{p(\mathbf{z}|\mathbf{x}) p(\mathbf{y}|\mathbf{z},\mathbf{x})}{q(\mathbf{z}|\mathbf{x},\mathbf{y},\mathbf{\boldsymbol \pi})} d\mathbf{z} \\
&=& \int q(\mathbf{z}|\mathbf{x},\mathbf{y},\mathbf{\boldsymbol \pi}) \left( \log \frac{p(\mathbf{z}|\mathbf{x})}{q(\mathbf{z}|\mathbf{x},\mathbf{y})} + \log p(\mathbf{y}|\mathbf{z},\mathbf{x}) \right ) d\mathbf{z} \\
&=& -\mathrm{D}_{\mathrm{KL}}\left[q(\mathbf{z}|\mathbf{x},\mathbf{y},\mathbf{\boldsymbol \pi}) || p(\mathbf{z}|\mathbf{x}) \right] + \mathbb{E}_{q(\mathbf{z}|\mathbf{x},\mathbf{y},\mathbf{\boldsymbol \pi})}\left[ \log p(\mathbf{y}|\mathbf{z},\mathbf{x}) \right]\\
&=& \mathcal{L}
\end{eqnarray*}

\section{Conditional GRU}

Conditional GRU is implemented in {\it dl4mt}.
\cite{Does_multi} extends Conditional GRU to make it capable of receiving image information as input. 
The first GRU computes intermediate representation $s'_j$ as
\begin{eqnarray}
\mathbf{s}'_j = (1 - \mathbf{o}'_j) \odot \underline{\mathbf{s}}'_j + \mathbf{o}'_j \odot \mathbf{s}_{j-1}\\
\underline{\mathbf{s}}'_j = \mathrm{tanh}(W'E\left[\mathbf{y}_{j-1}\right] + \mathbf{r}'_j \odot (U'\mathbf{s}_{j-1}))\\
\mathbf{r}'_j = \sigma(W'_r E\left[\mathbf{y}_{j-1}\right] + U'_r \mathbf{s}_{j-1})\\
\mathbf{o}'_j = \sigma(W'_o E\left[\mathbf{y}_{j-1}\right] + U'_o \mathbf{s}_{j-1})
\end{eqnarray}
where $E \in \mathbb{R}^ {d_{emb} \times d_t}$ signifies the target word embedding, $\underline{\mathbf{s}}'_j \in \mathbb{R}^ {d_h}$ denotes the hidden state, $\mathbf{r}'_j \in \mathbb{R}^ {d_h}$ and $\mathbf{o}'_j \in \mathbb{R}^ {d_h}$ respectively represent the reset and update gate activations. $d_t$ stands for the dimension of target; the unique number of target words. $\left[W', W'_r, W'_o \right] \in \mathbb{R}^ {d_h \times d_{emb}}, \left[U', U'_r, U'_o \right] \in \mathbb{R}^ {d_h \times d_h}$ are the parameters to be learned.

Context vector $\mathbf{c}_j$ is obtained as
\begin{eqnarray}
    \mathbf{c}_j = \mathrm{tanh}\left(\sum_{i=1}^{T_f} \alpha_{ij} \mathbf{h}_i\right) \label{eq:context}\\
 \alpha_{ij} =\frac{\exp(\mathbf{e}_{ij})}{\sum_{k=1}^{T_f} \exp(e_{kj})}\\
 \mathbf{e}_{ij} = U_{att} \mathrm{tanh}(W_{catt} \mathbf{h}_i + W_{att}\mathbf{s}'_j)
\end{eqnarray}
where \ $\left[U_{att}, W_{catt}, W_{att} \right] \in \mathbb{R}^ {d_h \times d_h}$ are the parameters to be learned.

The second GRU computes $\mathbf{s}_j$ from $\mathbf{s}'_j$, $\mathbf{c}_j$ and $\mathbf{h}'_e$ as
\begin{eqnarray}
\mathbf{s}_j = (1 - \mathbf{o}'_j) \odot \underline{\mathbf{s}}_j+ \mathbf{o}_j \odot \mathbf{s}'_{j}\\
\underline{\mathbf{s}}_j = \mathrm{tanh}(W\mathbf{c}_j + \mathbf{r}_j \odot (U\mathbf{s}'_{j}) + V \mathbf{h}'_e)\\
\mathbf{r}_j = \sigma(W_r \mathbf{c}_j+ U_r \mathbf{s}'_{j} + V_r \mathbf{h}'_e)\\
\mathbf{o}_j = \sigma(W_o \mathbf{c}_j+ U_o \mathbf{s}'_{j} + V_o \mathbf{h}'_e)
\end{eqnarray}
where $\underline{\mathbf{s}}_j \in \mathbb{R}^ {d_h}$ stands for the hidden state, $\mathbf{r}_j \in \mathbb{R}^ {d_h}$ and $\mathbf{o}_j \in \mathbb{R}^ {d_h}$ are the reset and update gate activations. $\left[W, W_r, W_o \right] \in \mathbb{R}^ {d_h \times d_h}, \left[U, U_r, U_o \right] \in \mathbb{R}^ {d_h \times d_h}, \left[V, V_r, V_o \right]\in \mathbb{R}^ {d_h \times d_z}$ are the parameters to be learned. We introduce $\mathbf{h}'_e$ obtained from a latent variable here so that a latent variable can affect the representation $\mathbf{s}_j$ through GRU units. 

Finally, the probability of $y$ is computed as
\begin{eqnarray}
    \mathbf{u}_j = L_u \mathrm{tanh} (E\left[\mathbf{y}_{j-1}\right] + L_s \mathbf{s}_j + L_x \mathbf{c}_j)\\
    P(\mathbf{y}_j | \mathbf{y}_{j-1}, \mathbf{s}_j,\mathbf{c}_j) = \mathrm{Softmax}(\mathbf{u}_j)
\end{eqnarray}
where $L_u \in \mathbb{R}^{d_t \times d_{emb}}$, $L_s \in \mathbb{R}^{d_{emb} \times d_h}$ and $L_c \in \mathbb{R}^{d_{emb} \times d_h}$ are the parameters to be learned.

\section{Training Detail}

\subsection{Hyperparameters}
Table \ref{table:param} presents parameters that we use in the experiments.

\begin{table}[htbp]
    \begin{center}
        \caption{Hyperparameters. The name is the variable name of {\it dl4mt} except for {\it dimv} and {\it dim\_pic}, which are the dimension of the latent variables and image embeddings.
        We set {\it dim} (number of LSTM unit size) and {\it dim\_word} (dimensions of word embeddings) 256, {\it batchsize} 32, {\it maxlen} (max output length) 50 and {\it lr} (learning rate) 1.0 for all models. {\it decay-c} is weights on L2 regularization.}
        \vspace{2mm}
        \label{table:param}
        \begin{tabular}{ll|l|l|l}
            &                 & {\it dimv} & {\it dim\_pic} & {\it decay-c} \\
                        \hline
                        & NMT             & -   &  256 &  0.001 \\
                        & VNMT            & 256 &  256 &  0.0005\\
            \hline
            Our Model   & G             & 256 &  512 &  0.001 \\
                        & G+O-AVG        & 256 &  256 &  0.0005 \\
                        & G+O-RNN        & 256 &  256 &  0.0005 \\
                        & G+O-TXT        & 256 &  256 &  0.0005 \\
        \end{tabular}
    \end{center}
\end{table}

\begin{figure}[htbp]
\begin{tabular}{c}
\begin{minipage}{0.46\hsize}
  \begin{center}
      \includegraphics[scale=0.33]{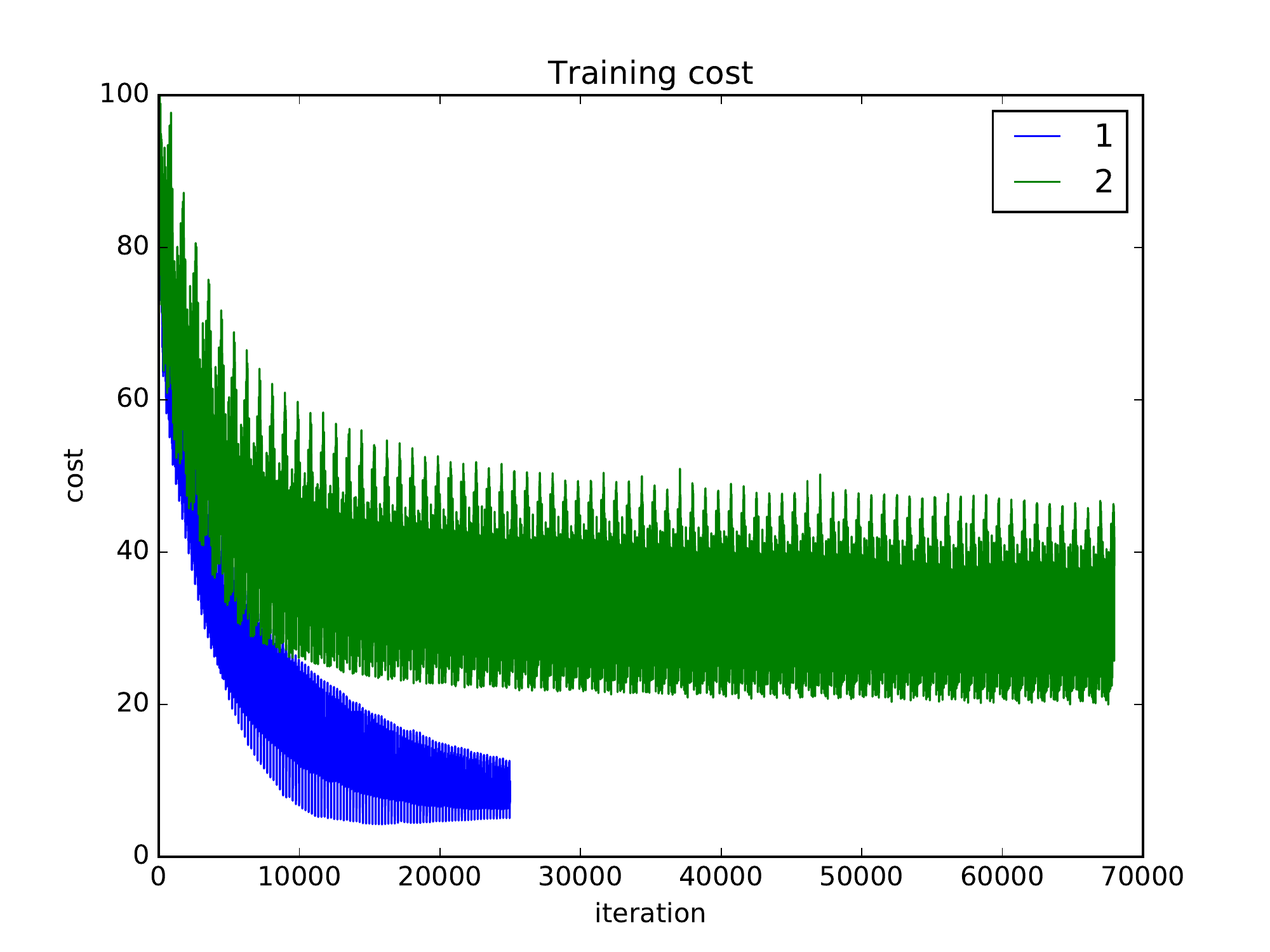}
      \caption{NMT Training Cost}
    \label{fig:nmt_cost_compare}
  \end{center}
\end{minipage}
\hspace{5mm}

\begin{minipage}{0.46\hsize}
  \begin{center}
      \includegraphics[scale=0.33]{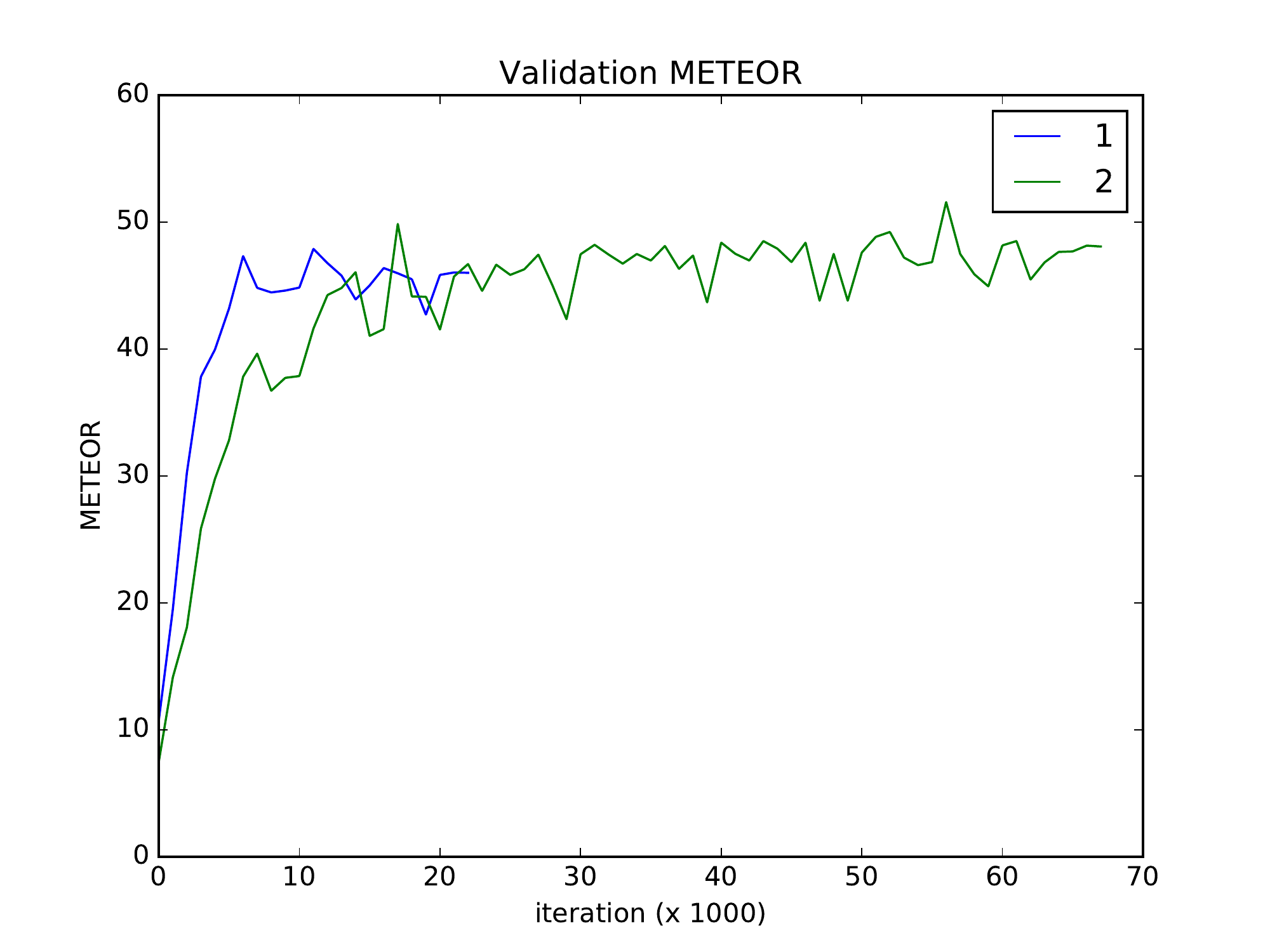}
      \caption{NMT Validation METEOR score}
    \label{fig:nmt_meteor_compare}
  \end{center}
\end{minipage}
\end{tabular}
\end{figure}

We found that Multi30k dataset is easy to overfit. Figure \ref{fig:nmt_cost_compare} and Figure \ref{fig:nmt_meteor_compare} present training cost and validation METEOR score graph of the two experimental settings of the NMT model. Table \ref{table:nmt_compare} presents the hyperparameters which were used in the experiments. Large {\it decay-c} ans small {\it batchsize} give the better METEOR scores in the end. Training is stopped if there is no validation cost improvements over the last 10 validations.

\begin{table}[htbp]
    \begin{center}
        \caption{Hyperparameters using the experiments in the Figure \ref{fig:nmt_cost_compare} and \ref{fig:nmt_meteor_compare}}
        \vspace{2mm}
        \label{table:nmt_compare}
        \begin{tabular}{l|llllll}
                         &  {\it dim} & {\it dim\_word} & {\it lr} & {\it decay-c} & {\it maxlen} & {\it batchsie} \\ \hline
            1    & 256        & 256             &  1.0 &  0.0005 & 30 & 128 \\
            2    & 256        & 256             &  1.0 &  0.001 & 50 & 32 \\
        \end{tabular}
    \end{center}
\end{table}

\begin{figure}[htbp]
  \begin{center}
      \includegraphics[scale=0.35]{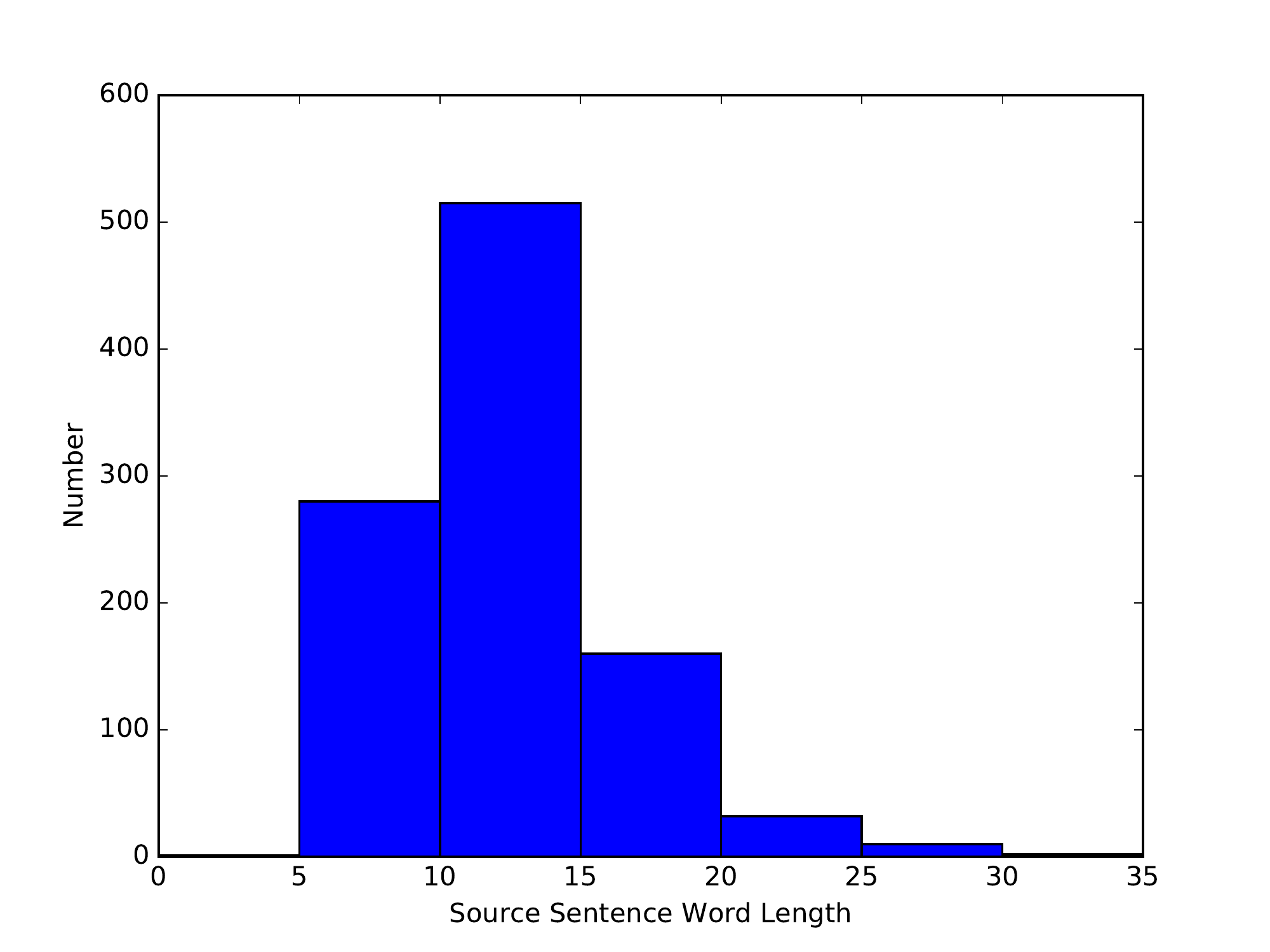}
      \caption{Word Length Histogram of the Multi30k Test Dataset}
    \label{fig:length_hist}
  \end{center}
\end{figure}

Figure \ref{fig:length_hist} presents the English word length histogram of the Multi30k test dataset. Most sentences in the Multi30k are less than 20 words. We assume that this is one of the reasons why Multi30k is easy to overfit.

\subsection{cost graph}
\begin{figure}[htbp]
    \begin{center}
        \begin{tabular}{ccc}
            \subfigure[NMT]{ \includegraphics[width=45mm] {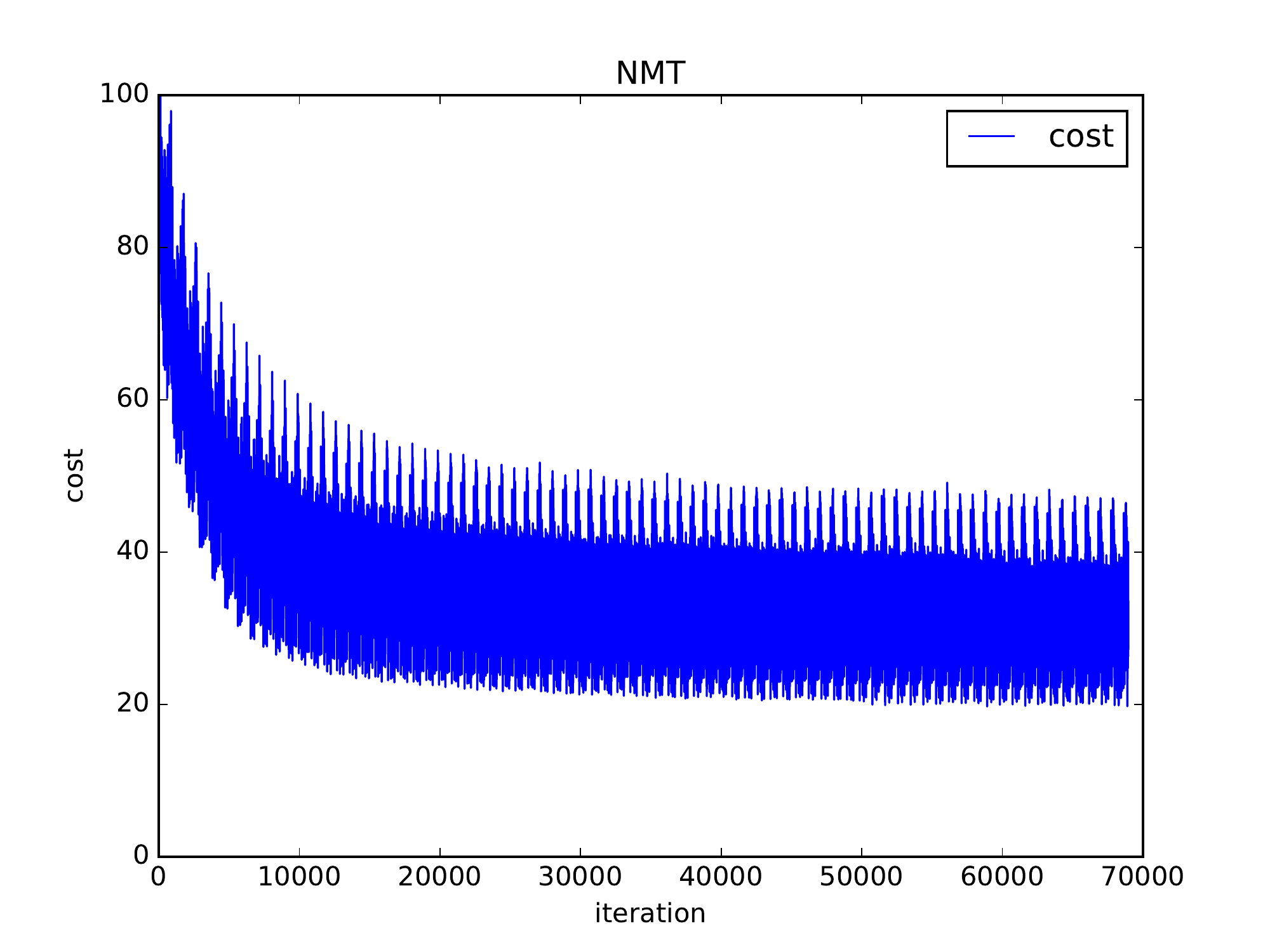}} &
            \subfigure[VNMT]{ \includegraphics[width=45mm] {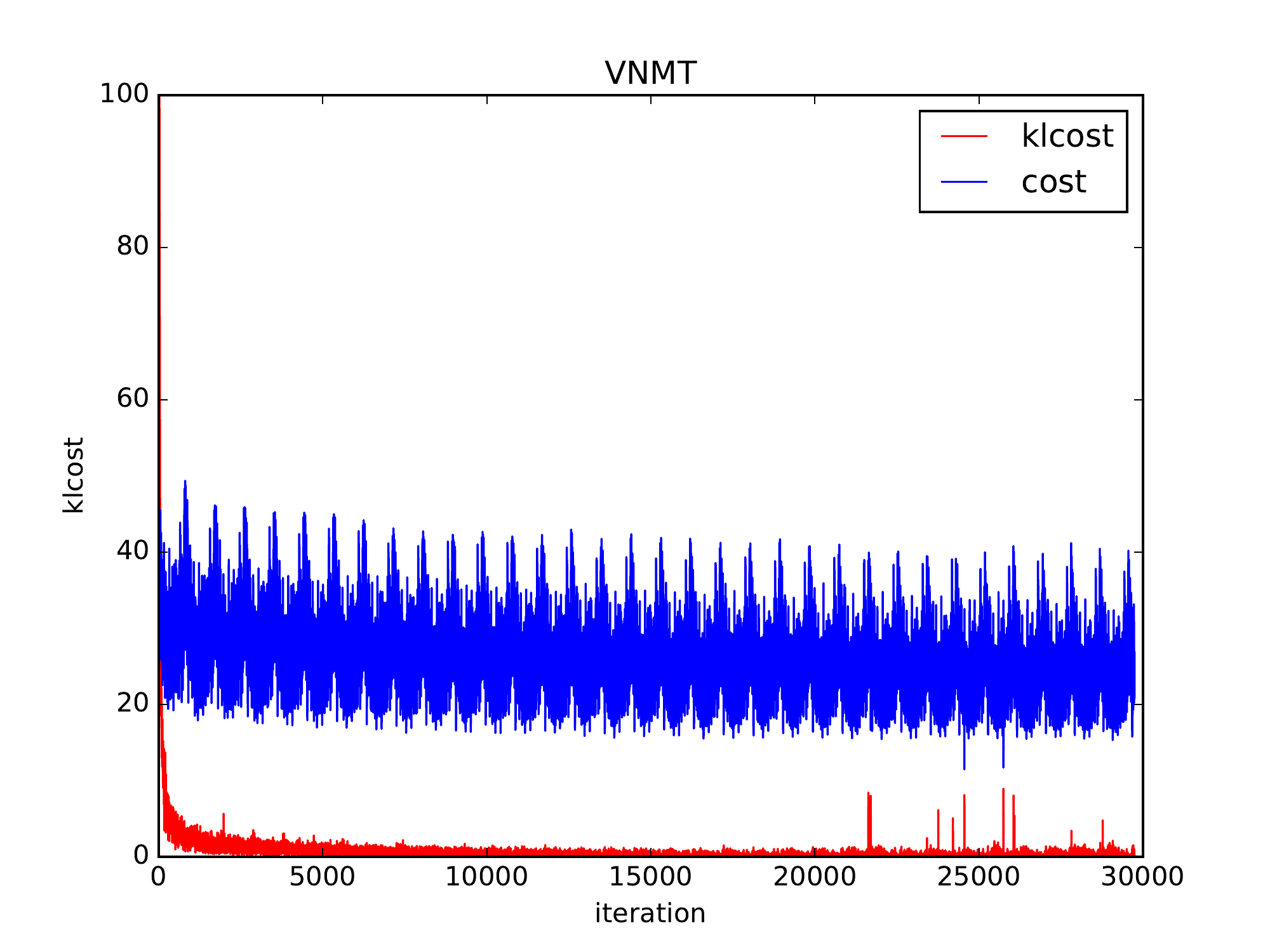}} &
            \subfigure[G]{ \includegraphics[width=45mm] {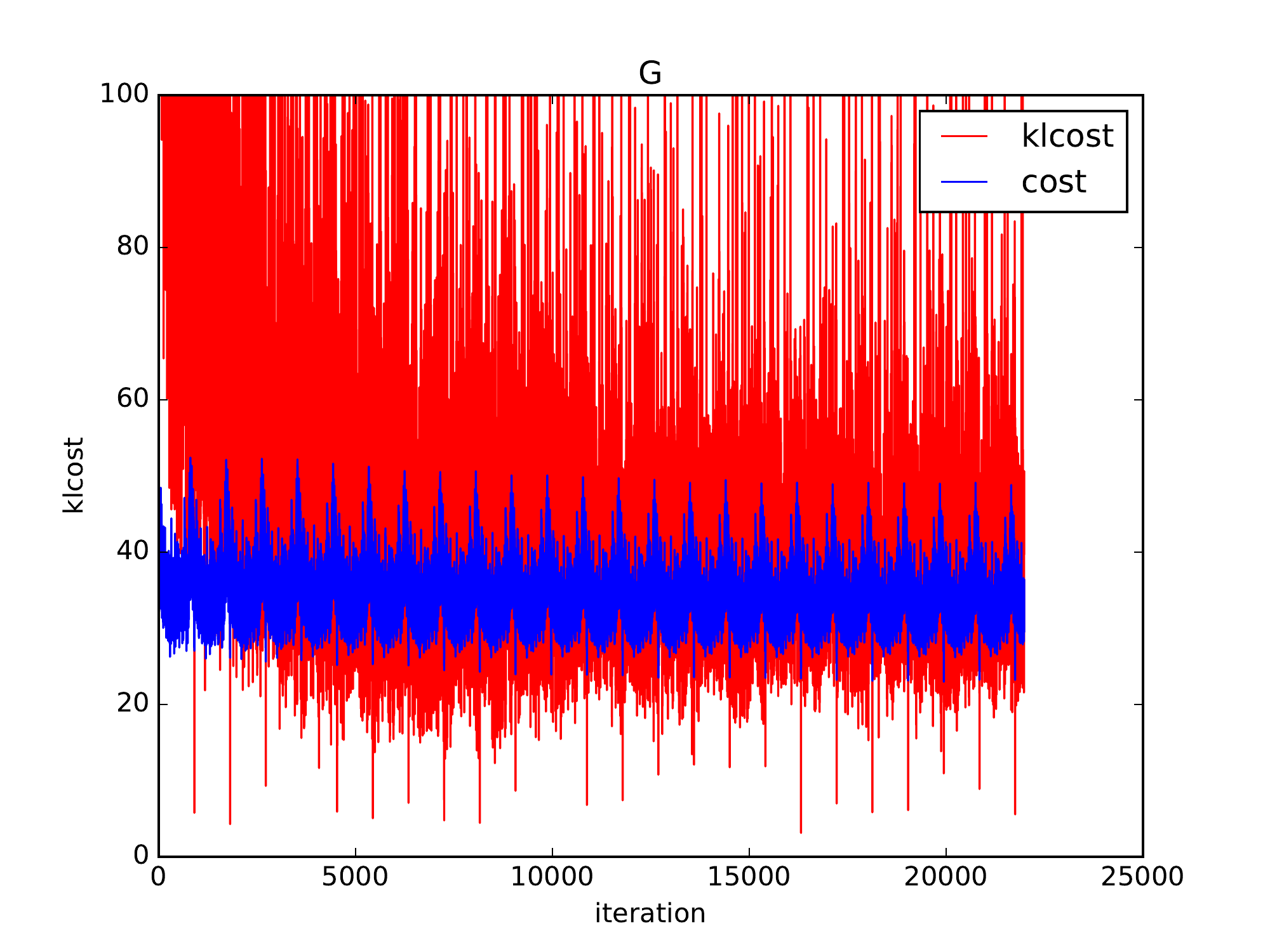}} \\
            \subfigure[G+O-AVG]{ \includegraphics[width=45mm] {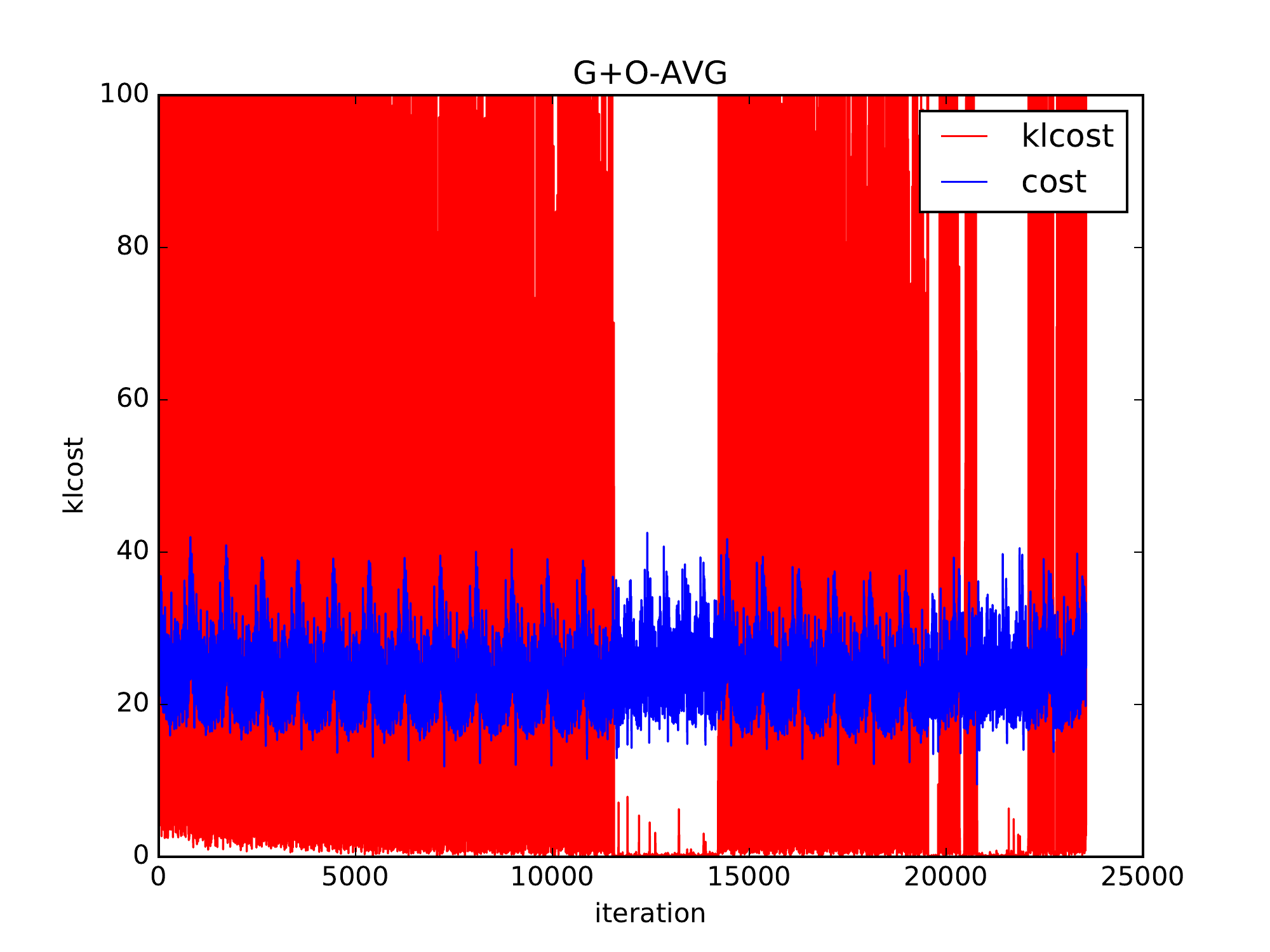}} &
            \subfigure[G+O-RNN]{ \includegraphics[width=45mm] {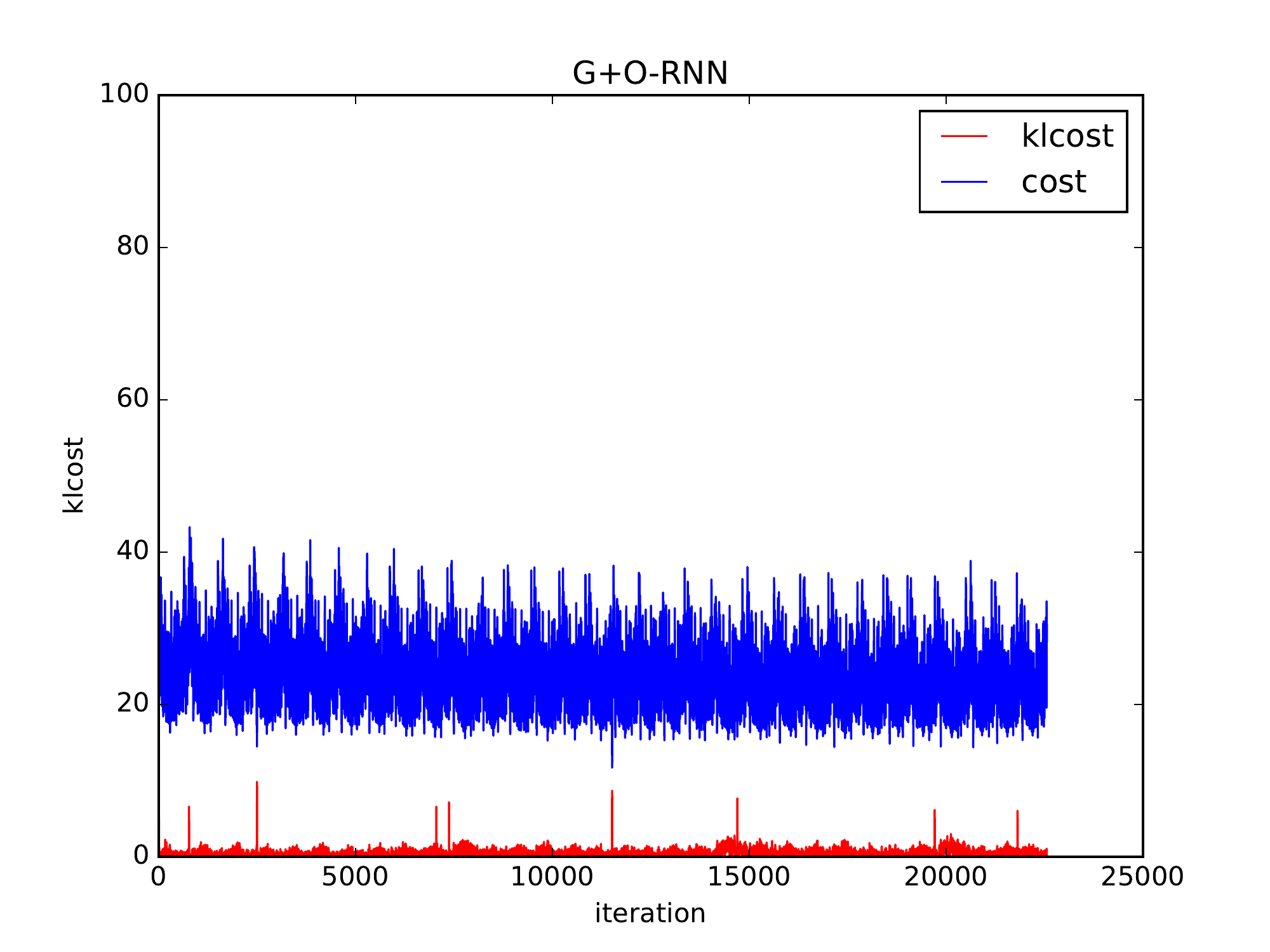}} &
            \subfigure[G+O-TXT]{ \includegraphics[width=45mm] {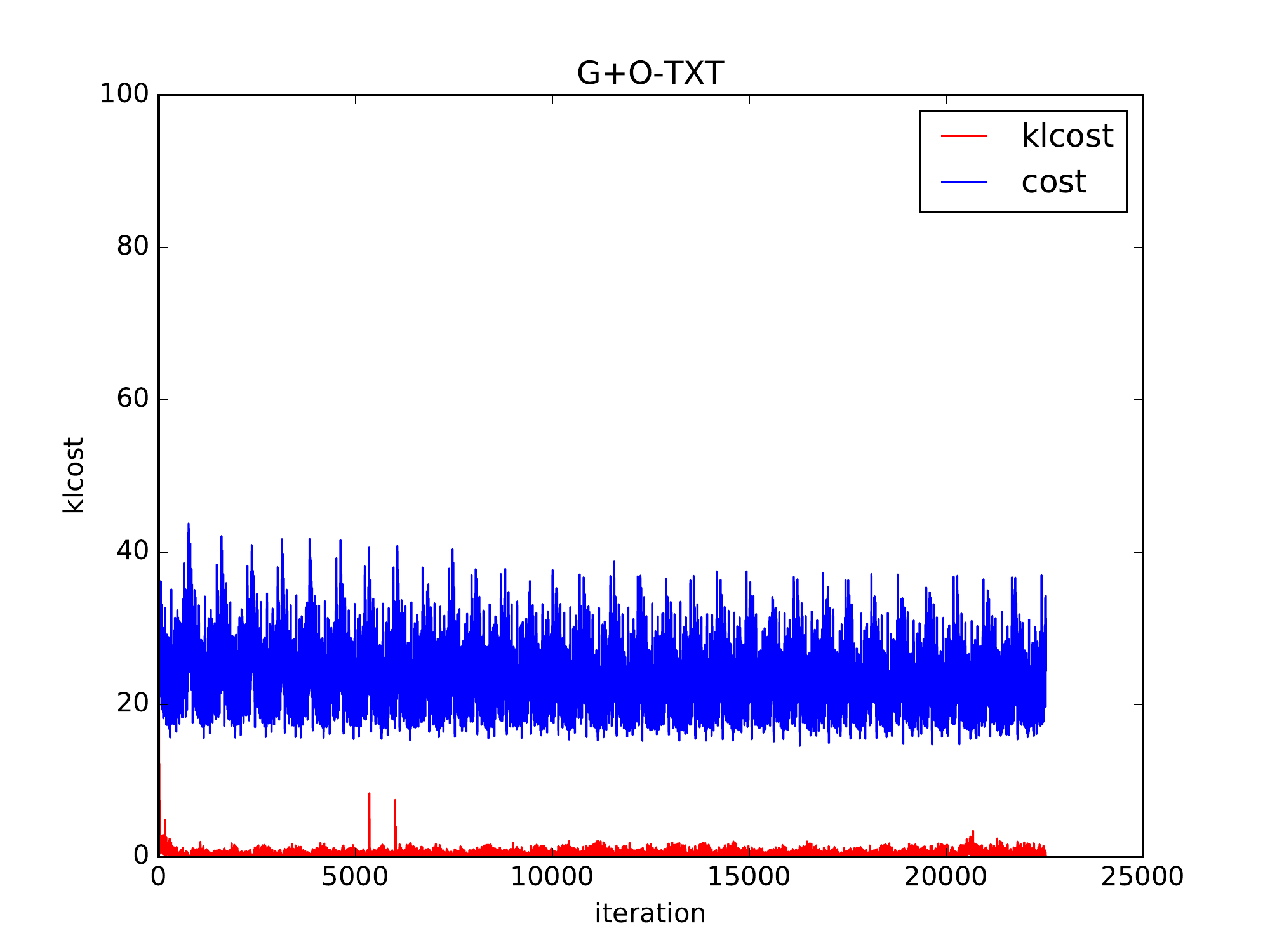}} \\
    \end{tabular}
    \end{center}
    \caption{Training cost}
    \label{fig:cost}
\end{figure}

\begin{figure}[htbp]
    \begin{center}
        \begin{tabular}{ccc}
            \subfigure[NMT]{ \includegraphics[width=45mm] {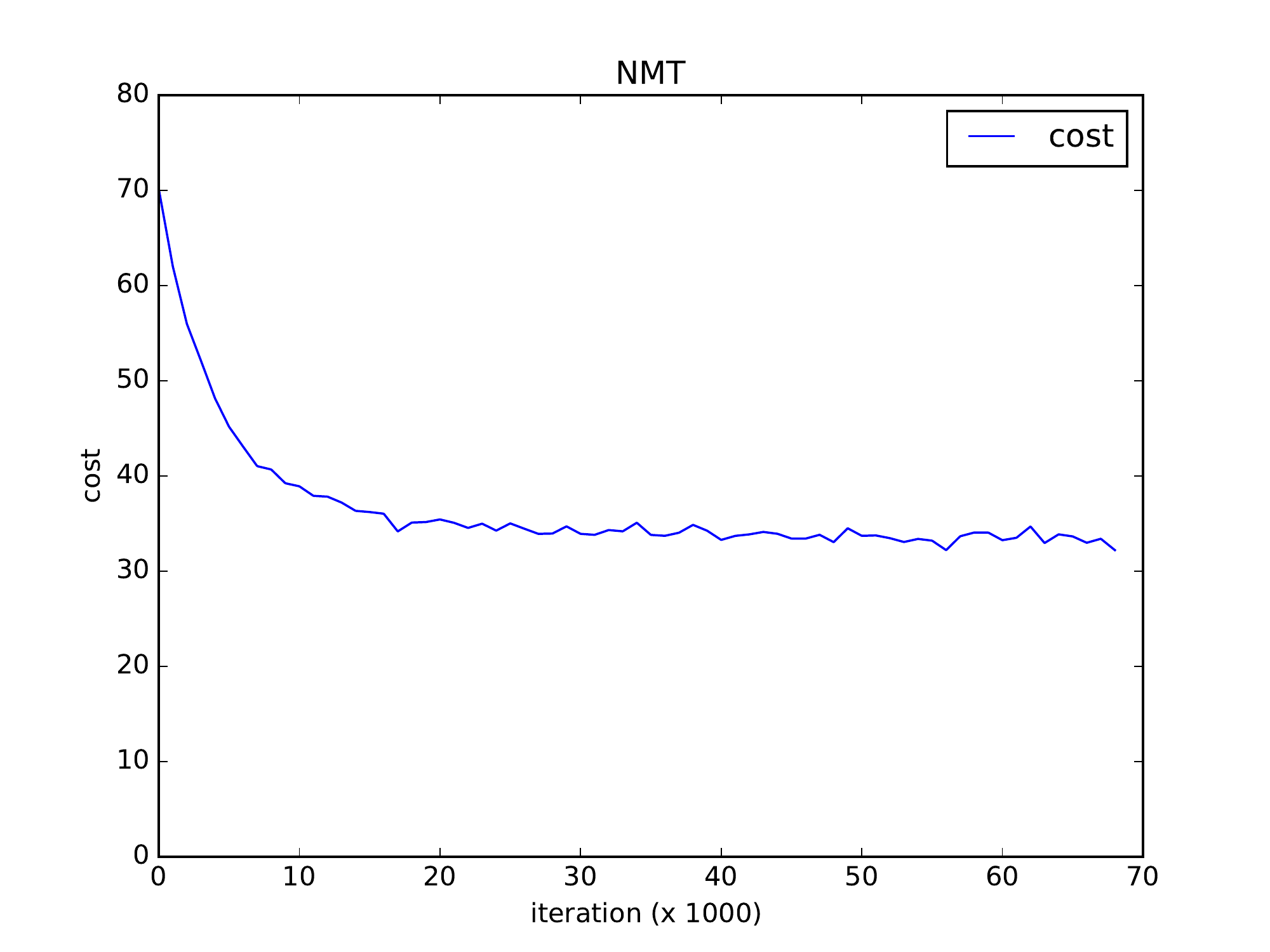}} &
            \subfigure[VNMT]{ \includegraphics[width=45mm] {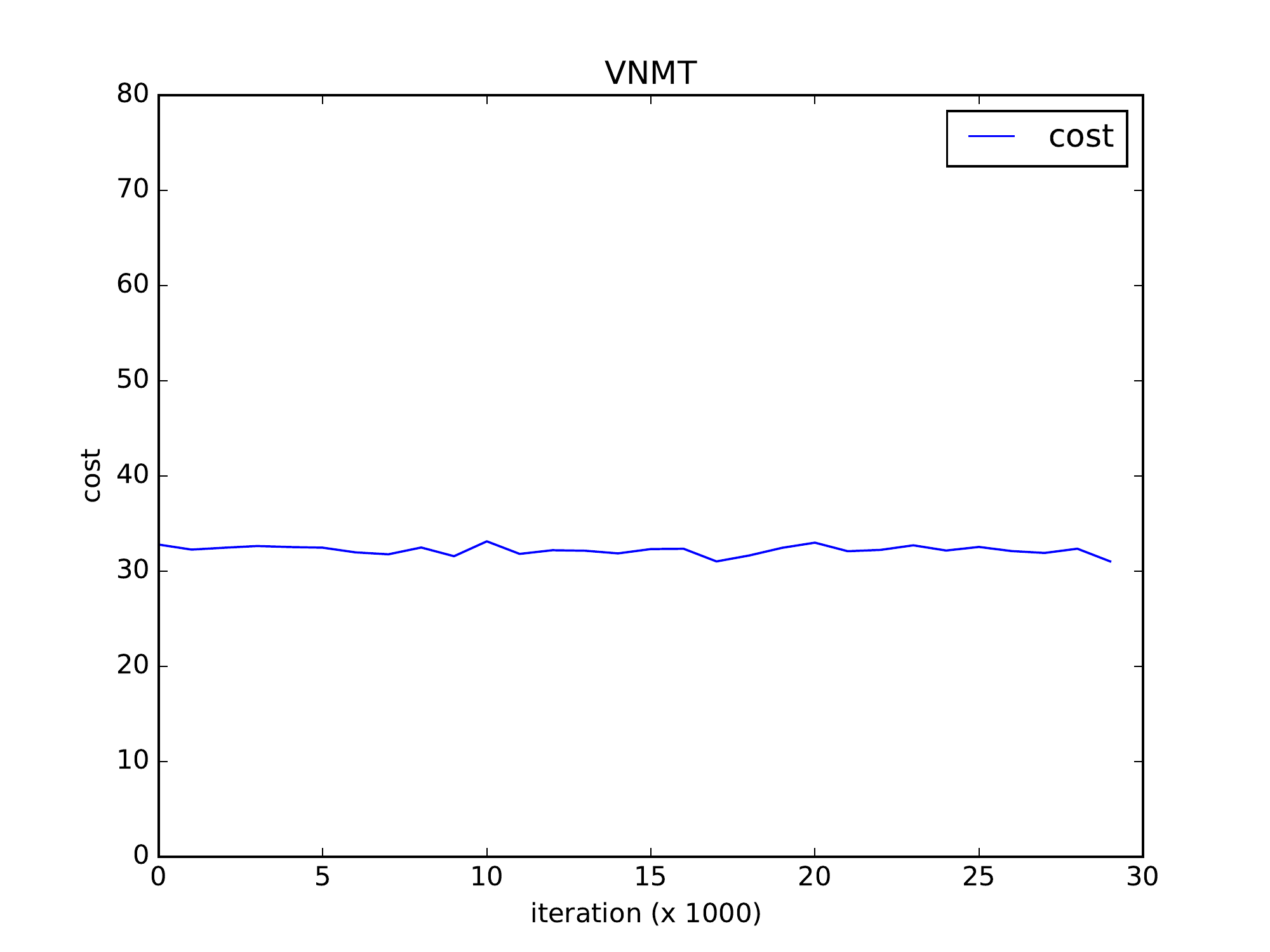}} &
            \subfigure[G]{ \includegraphics[width=45mm] {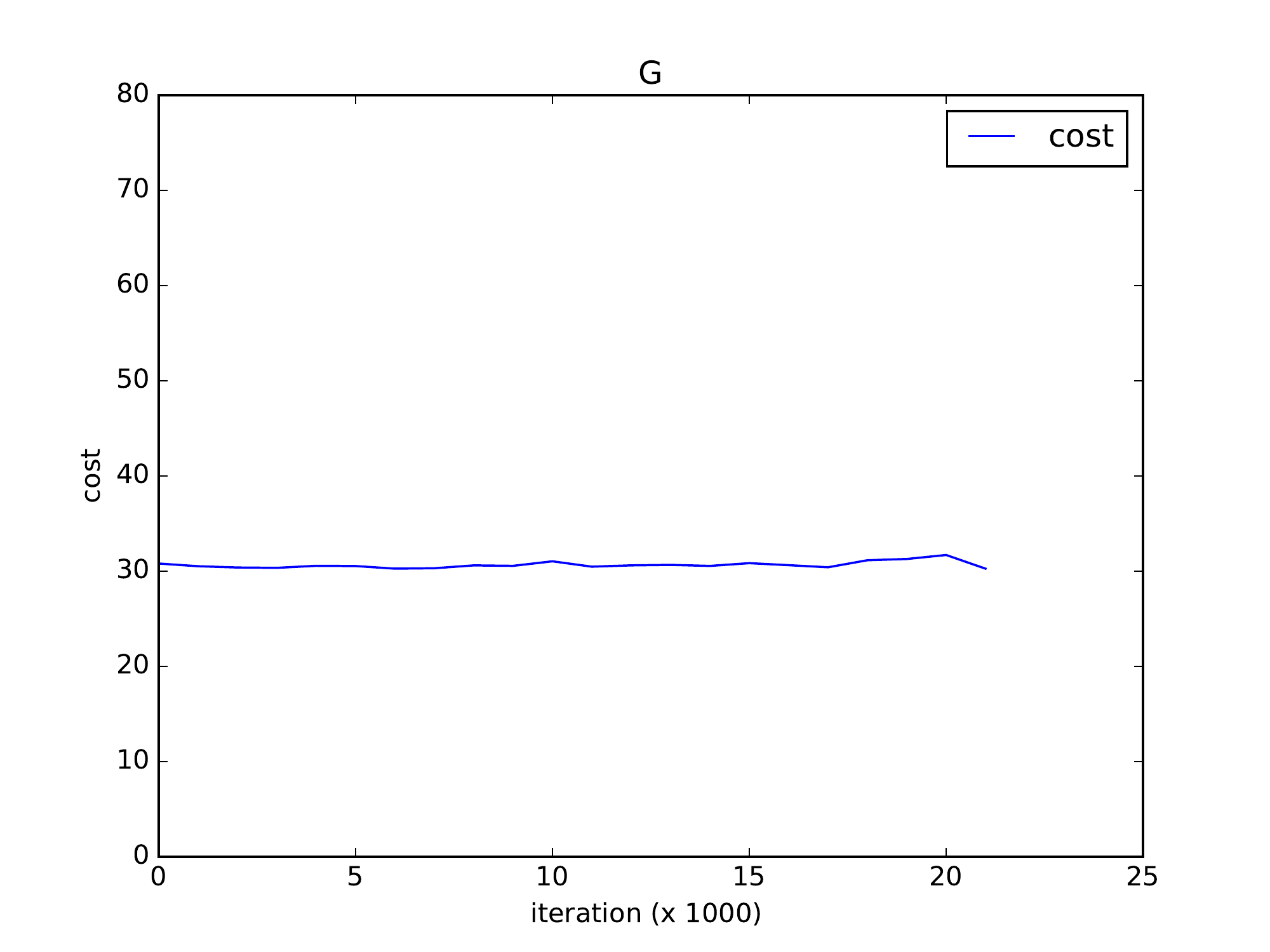}} \\
            \subfigure[G+O-AVG]{ \includegraphics[width=45mm] {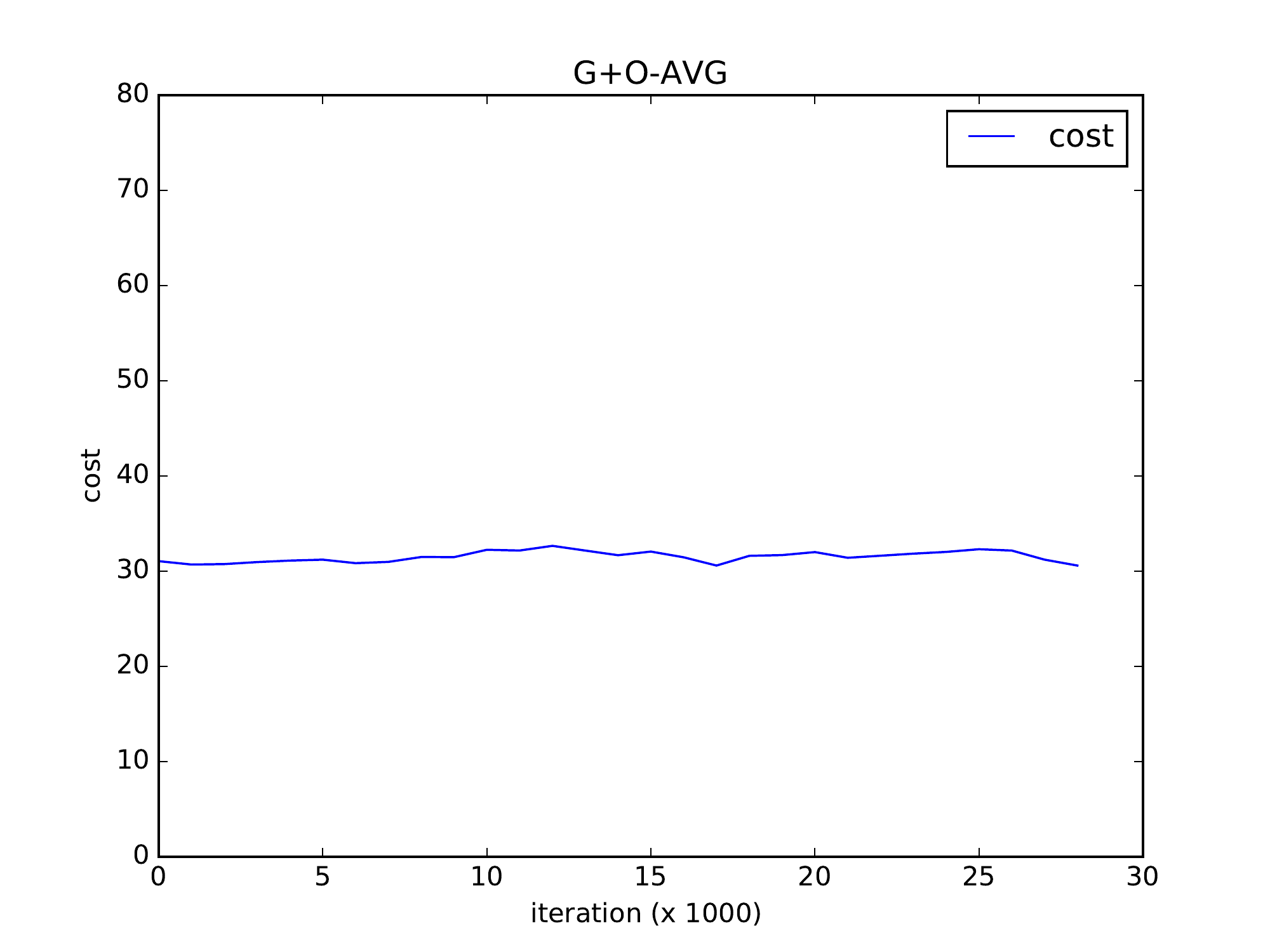}} &
            \subfigure[G+O-RNN]{ \includegraphics[width=45mm] {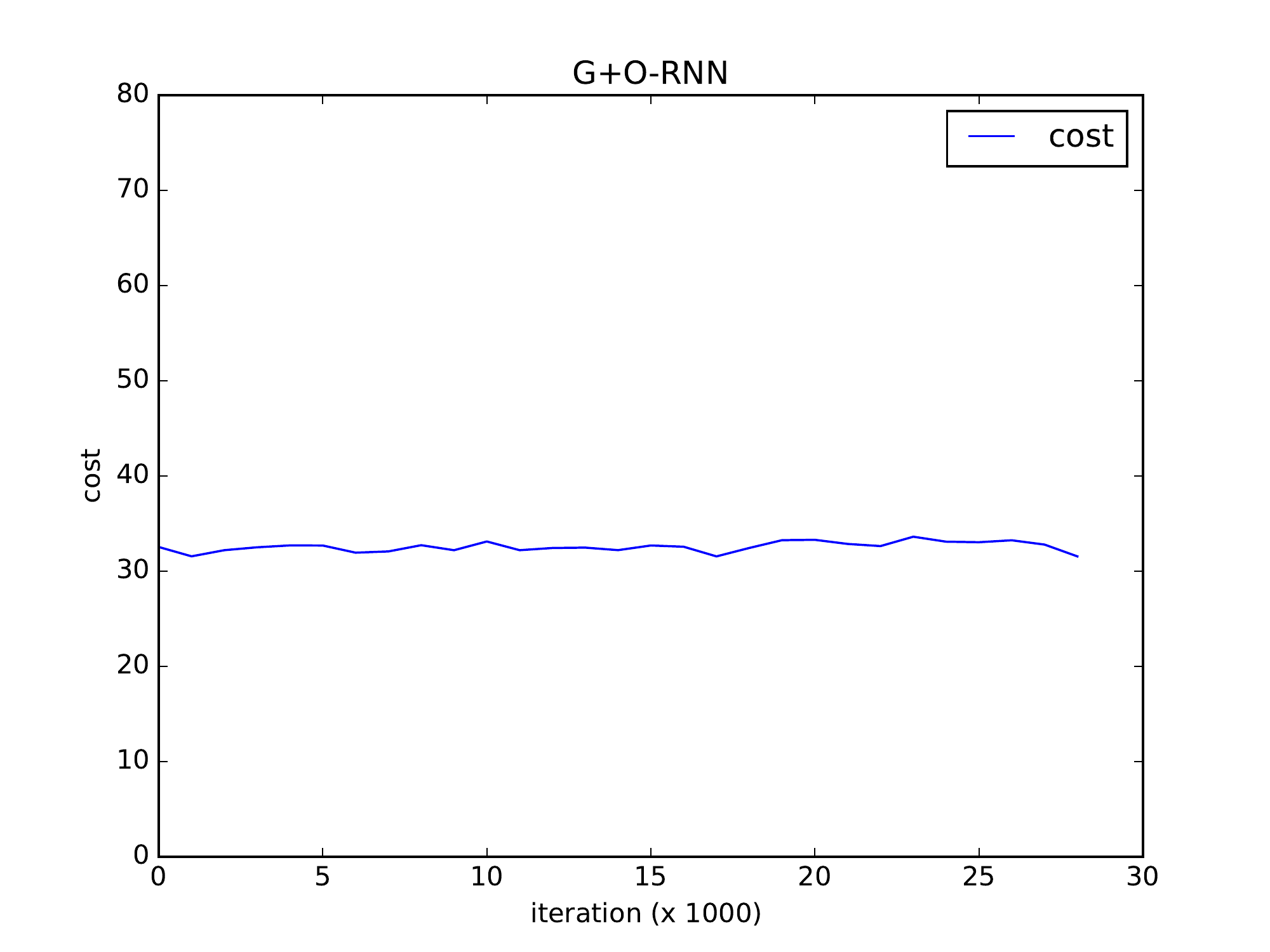}} &
            \subfigure[G+O-TXT]{ \includegraphics[width=45mm] {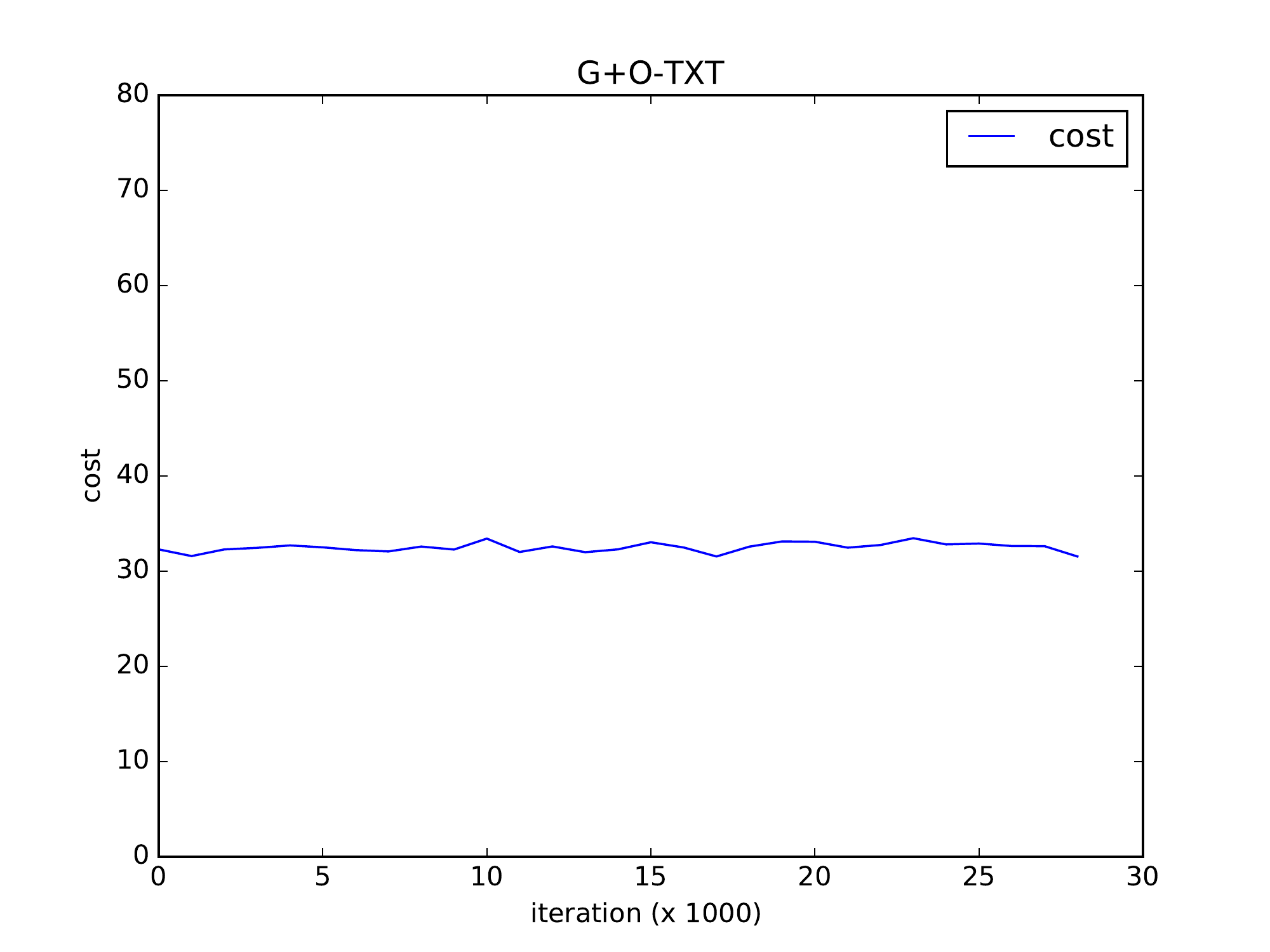}} \\
    \end{tabular}
    \end{center}
    \caption{Validation cost}
    \label{fig:valcost}
\end{figure}

Figure \ref{fig:cost} and \ref{fig:valcost} present the training cost and validation cost graph of each models. Please note that VNMT fine-tuned NMT, and other models fine-tuned VNMT.

\subsection{Translation Examples}
We present some selected translations from VNMT and our proposed model (G). As of translation 3 to 5 our model give the better METEOR scores than VNMT and as of translation 6 to 8 VNMT give the better METEOR scores than our models.


\begin{table}[htbp]
    \begin{center}
        \begin{tabular}{|l|l|}
            \multicolumn{2}{c}{\includegraphics[scale=0.5]{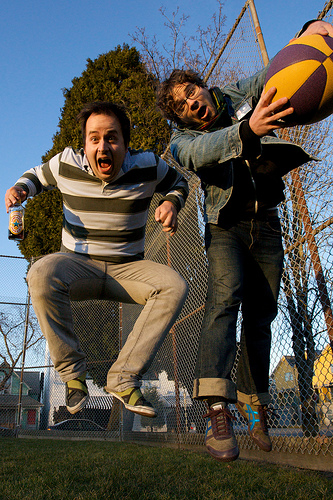}}\\
            \hline
            {\bf Source}  & 
            two boys inside a fence jump in the air while holding a basketball. \\ \hline
            {\bf Truth}  &
            zwei jungen innerhalb eines zaunes springen in die luft und halten dabei einen basketball. \\ \hline
            {\bf VNMT} &
            zwei jungen in einem zaun springen in die luft, w\"ahrend sie einen basketball h\"alt. \\ \hline
            {\bf Our Model (G)} &
            zwei jungen in einem zaun springen in die luft und halten dabei einen basketball. \\ \hline
        \end{tabular}
        \figcaption{Translation 3}
    \end{center}
\end{table}

\begin{table}[htbp]
    \begin{center}
        \begin{tabular}{|l|l|}
            \multicolumn{2}{c}{\includegraphics[scale=0.5]{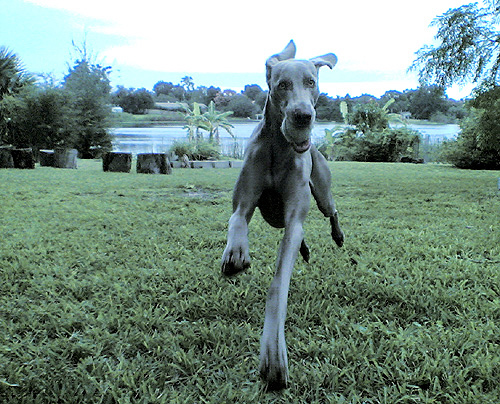}}\\
            \hline
            {\bf Source}  & 
            a dog runs through the grass towards the camera.
            \\ \hline
            {\bf Truth}  &
            ein hund rennt durch das gras auf die kamera zu.
            \\ \hline
            {\bf VNMT} &
            ein hund rennt durch das gras in die kamera.
            \\ \hline
            {\bf Our Model (G)} &
            ein hund rennt durch das gras auf die kamera zu.
            \\ \hline
        \end{tabular}
        \figcaption{Translation 4}
    \end{center}
\end{table}

\begin{table}[htbp]
    \begin{center}
        \begin{tabular}{|l|l|}
            \multicolumn{2}{c}{\includegraphics[scale=0.5]{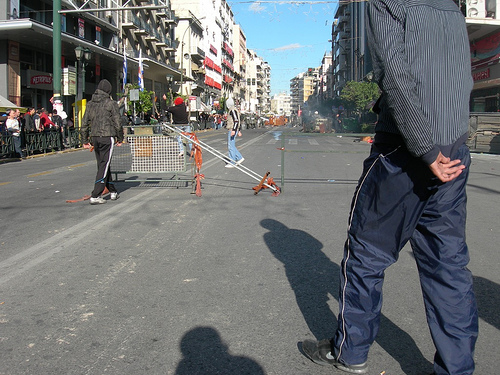}}\\
            \hline
            {\bf Source}  & 
            a couple of men walking on a public city street.
            \\ \hline
            {\bf Truth}  &
            einige m\"anner gehen auf einer \"offentlichen stra\ss e in der stadt.
            \\ \hline
            {\bf VNMT} &
            ein paar m\"anner gehen auf einer \"offentlichen stadtstra\ss e.
            \\ \hline
            {\bf Our Model (G)} &
            ein paar m\"anner gehen auf einer \"offentlichen stra\ss e in der stadt.
            \\ \hline
        \end{tabular}
        \figcaption{Translation 5}
    \end{center}
\end{table}

\begin{table}[htbp]
    \begin{center}
        \begin{tabular}{|l|l|}
            \multicolumn{2}{c}{\includegraphics[scale=0.5]{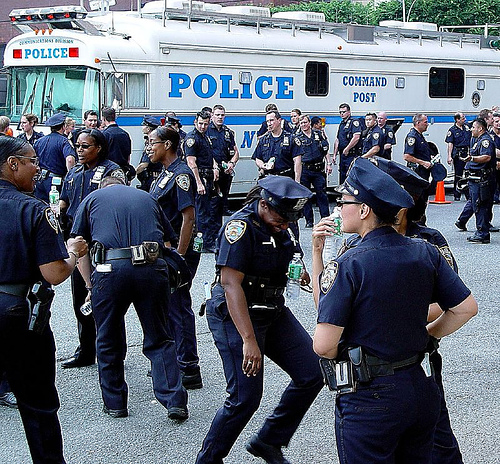}}\\
            \hline
            {\bf Source}  &
            a bunch of police officers are standing outside a bus.
            \\ \hline
            {\bf Truth}  &
            eine gruppe von polizisten steht vor einem bus.
            \\ \hline
            {\bf VNMT} &
            eine gruppe von polizisten steht vor einem bus.
            \\ \hline
            {\bf Our Model (G)} &
            mehrere polizisten stehen vor einem bus.
            \\ \hline
        \end{tabular}
        \figcaption{Translation 6}
    \end{center}
\end{table}

\begin{table}[htbp]
    \begin{center}
        \begin{tabular}{|l|l|}
            \multicolumn{2}{c}{\includegraphics[scale=0.5]{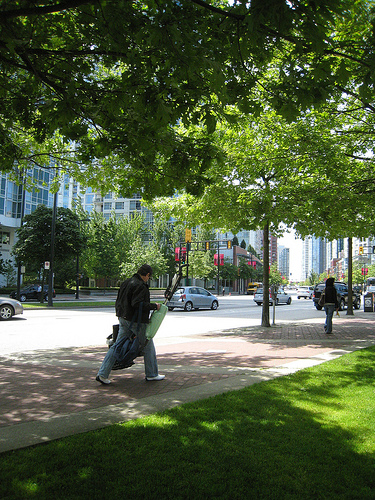}}\\
            \hline
            {\bf Source}  &
            a man is walking down the sidewalk next to a street.
            \\ \hline
            {\bf Truth}  &
            ein mann geht neben einer stra\ss e den gehweg entlang.
            \\ \hline
            {\bf VNMT} &
            ein mann geht neben einer stra\ss e den b\"urgersteig entlang.
            \\ \hline
            {\bf Our Model (G)} &
            ein mann geht auf dem b\"urgersteig an einer stra\ss e.
            \\ \hline
        \end{tabular}
        \figcaption{Translation 7}
    \end{center}
\end{table}

\begin{table}[htbp]
    \begin{center}
        \begin{tabular}{|l|l|}
            \multicolumn{2}{c}{\includegraphics[scale=0.5]{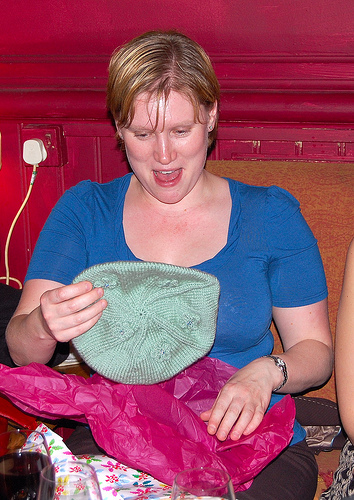}}\\
            \hline
            {\bf Source}  &
            a blond-haired woman wearing a blue shirt unwraps a hat.
            \\ \hline
            {\bf Truth}  &
            eine blonde frau in einem blauen t-shirt packt eine m\"utze aus.
            \\ \hline
            {\bf VNMT} &
            eine blonde frau in einem blauen t-shirt wirft einen hut.
            \\ \hline
            {\bf Our Model (G)} &
            eine blonde frau tr\"agt ein blaues hemd und einen hut.
            \\ \hline
        \end{tabular} 
        \figcaption{Translation 8}
    \end{center}
\end{table}

\end{document}